\documentclass[letterpaper]{article} 
\usepackage{aaai25}  
\usepackage{times}  
\usepackage{helvet}  
\usepackage{courier}  
\usepackage[hyphens]{url}  
\usepackage{graphicx} 
\usepackage{natbib}  
\usepackage{caption} 
\usepackage{algorithm}
\usepackage{algorithmic}
\usepackage{amssymb}
\usepackage{multirow}
\usepackage{booktabs}
\usepackage{amsmath}
\usepackage{bbding}
\usepackage{cleveref}
\usepackage{newfloat}
\usepackage{listings}
\DeclareCaptionStyle{ruled}{labelfont=normalfont,labelsep=colon,strut=off} 
\floatstyle{ruled}
\newfloat{listing}{tb}{lst}{}
\floatname{listing}{Listing}
\title{Multi-Pair Temporal Sentence Grounding via Multi-Thread Knowledge Transfer Network}
\author{
    Xiang Fang\textsuperscript{\rm 1,2}, Wanlong Fang\textsuperscript{\rm 3}, Changshuo Wang\textsuperscript{\rm 3}\thanks{Corresponding Author.}, 
    Daizong Liu\textsuperscript{\rm 4}, Keke Tang\textsuperscript{\rm 5}, Jianfeng Dong\textsuperscript{\rm 6}, Pan Zhou\textsuperscript{\rm 1}$^{*}$, Beibei Li\textsuperscript{\rm 2}$^{*}$
}
\affiliations{
    \textsuperscript{\rm 1}Hubei Engineering Research Center on Big Data Security, School of Cyber Science \\and Engineering, Huazhong University of Science of Technology Wuhan, China\\ 
    \textsuperscript{\rm 2}Sichuan University \\
    \textsuperscript{\rm 3}Nanyang Technological University, Singapore \\
    \textsuperscript{\rm 4}Peking University \\
    \textsuperscript{\rm 5}Guangzhou University \\
    \textsuperscript{\rm 6}Zhejiang Gongshang University 

     xfang9508@gmail.com,  wanlongfang@gmail.com, wangchangshuo1@gmail.com, dzliu@stu.pku.edu.cn, tangbohutbh@gmail.com, dongjf24@gmail.com, panzhou@hust.edu.cn, libeibei@scu.edu.cn
}
\usepackage{bibentry}

\begin{document}

 \maketitle

\begin{abstract}
Given some video-query pairs with  untrimmed videos and  sentence queries, temporal sentence grounding (TSG) aims to locate  query-relevant segments in these videos.
Although previous respectable TSG methods have achieved remarkable success, they  train each video-query pair separately and ignore the relationship between different pairs.
We observe that the similar video/query content not only helps the TSG model better understand and generalize the cross-modal representation, but also assists the model in locating some complex video-query pairs.
Previous methods follow a single-thread framework that cannot co-train different pairs and usually spends much time re-obtaining redundant knowledge, limiting their real-world applications.
To this end, in this paper,  we pose a brand-new setting: Multi-Pair TSG, which aims to co-train these pairs.
In particular, we propose a novel video-query co-training approach, Multi-Thread Knowledge Transfer Network, to  locate a variety of video-query pairs effectively and efficiently. 
Firstly, we  mine the spatial and temporal semantics across different queries to cooperate with each other. To learn  intra- and inter-modal representations simultaneously, we design a cross-modal contrast module to explore the semantic consistency by a self-supervised strategy. To fully align visual  and textual representations between different pairs, we design a prototype alignment strategy to 1) match  object prototypes and phrase prototypes for spatial alignment, and 2) align activity prototypes and sentence prototypes for temporal alignment. Finally, we develop an adaptive negative selection module to adaptively generate a threshold for cross-modal matching.
Extensive experiments show
the effectiveness and efficiency of our proposed method. 
\end{abstract}

%

\section{Introduction}\label{sec:intro}
\begin{figure}[t!]
    \centering
    \includegraphics[width=0.48\textwidth]{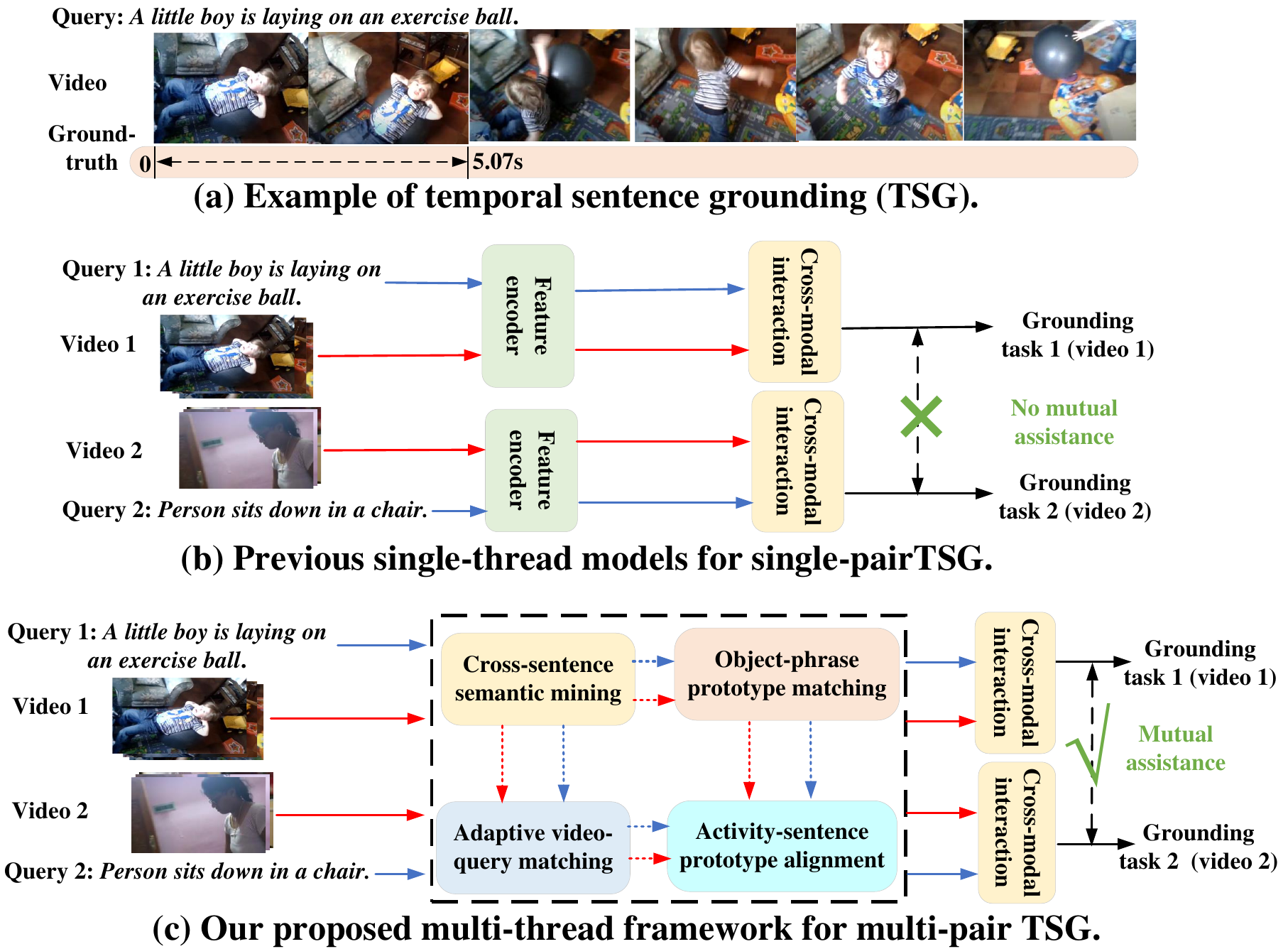}
     \vspace{-20pt}
    \caption{\small (a) Example of  temporal sentence grounding (TSG). (b) Previous TSG models regard each video-query pair independently. (c) Our proposed model explores the semantic relationship between different pairs to reduce the modality gap.}
\vspace{-23pt}
    \label{fig:intro}
\end{figure}

Temporal sentence grounding (TSG) \cite{gao2017tall,li2023event,li2024triplet,yu2024pedestrian,ning2024enhancement,ning2023occluded,ning2023pedestrian,wang2025looking} is an important yet challenging multi-modal task, which has received increasing attention in recent years due to its wide potential applications, such as video understanding \cite{liu2024survey,wang2024muki,hu2023password,fei2024enhancing,fei2024vitron,fei2024video,fei2024dysen,wu24next,wang2024gpsformer,wang20233d,wangchangshuo20223d,zhang2024deformation,zhang2024pointgt,YU2022108685,yu2025eds,liu2023exploring,wang2025taylor,fang2026towardsicml,kuai2026dynamic,wang2025point,fang2025your,zhang2025monoattack,fang2023hierarchical,liu2024towards,yang2025eood,fang2022multi,fang2026cogniVerse,lei2025exploring,fang2023you,wang2025dypolyseg,fang2025hierarchical,yan2026fit,fang2025adaptive,wang2026topadapter,cai2025imperceptible,fang2026slap,wang2026reasoning,fang2026immuno,wang2026biologically,fang2026disentangling,wang2025reducing,fang2026advancing,fang2026unveiling,wang2026from,liu2023conditional,liu2026attacking,fang2026rethinking,wang2025seeing,fang2026towards,fang2020double,fang2024fewer,liu2024pandora,fang2024multi,fang2025turing,fang2024not,liu2023hypotheses,fang2024rethinking,liu2024unsupervised,fang2023annotations,xiong2024rethinking,fang2021unbalanced,wang2025prototype,zhang2025manipulating,fang2026align,tang2024reparameterization,fang2025adaptivetai,tang2025simplification,fang2021animc,cai2026towards,fang2020v} and human-computer interaction \cite{liu2024a,liu2024pandora,liu2023spts,liu2023exploring,liu2023hypotheses,tang2022you,tang2022optimal,tang2023character,tang2024textsquare,feng2023unidoc,feng2023docpedia,zhao2024multi,zhao2024harmonizing,wang2024pargobridgingvisionlanguagepartial,10296854,9928031}.
By complex multi-modal interactions and complicated context information, TSG targets the challenging problem of locating a variety of sentence queries about a video, which requires the designed models to understand both natural language and long video, including reasoning about activities, objects, sequence of events, and interactions within the video \cite{HuHWJM22,WeiHZH23,zhao2021brain,zhao2017hdskg,zhao2022heterogeneous,zhao2018smart,zhao2018gamified,jia2019comdefend,jia2022adversarial,jia2022boosting,jia2022prior,jia2024revisiting,jia2021effective,jia2024fast,jia2020adv,jia2023transegpgd,jia2024semantic,jia2024improved,jia2024global,gao2021towards,gao2022two,gao2024dmofc,gao2024imofc,gao2024rethinking,fang2025adaptive,hu2020resilient,hu2021intrusion,hu2023resilient,hu2023resilient,hu2021credibility,hu2020distributed,hu2024robust,hu2020credibility,hu2024robust,hu2023security,hu2024resilient,hu2024general,hu2021resilient,hu2024novel,hu2024resilient,hu2024enhancing,hu2024resilient,hu2023abnormal,hu2023distributed,hu2023cyberattack,li2022numerical,li2025discrete}. 
As shown in Figure \ref{fig:intro}(a), given an untrimmed video and a sentence query, TSG aims to determine the segment boundaries that contain the query-relevant activity \cite{qu2020fine,dong2022reading,sun2024surf,qu2024alleviating,zheng2023progressive,qu2024look}. 


Most previous TSG works \cite{xiu2024hierarchical,liu2022few,liu2020saanet,liu2021spatiotemporal,ji2023binary,ji2023partial,ji2024weakly,ji2024described,ji2024backpropagation,ji2023towards} refer to a fully-supervised setting, where each frame is manually labeled as query-relevant or not.
To avoid using such labour-intensive frame-level annotations, some recent works \cite{fang2025rethinking,fang2020v,fang2020double,fang2021unbalanced,fang2021animc,fang2024uncertainty,fang2024your,liu2024moe,liu2023diffusion,liu2024multimodal,liullm,zhang2024less,yu2024recent,zhang2024geometric,zhang2023mitigating,zhang2023spectral,zhang2023contrastive,zhang2022costa,guo2024benchmarking,dong2022partially} explore a weakly-supervised setting with only the video-query correspondence to alleviate the reliance to a certain extent.
Despite the remarkable performance, fully- and weakly-supervised methods only treat each video-query pair independently and ignore the semantic relationship between different video-query pairs,  as shown in Figure \ref{fig:intro}(b). Since a video often corresponds to multiple queries, their treatment will repeatedly extract video features and repetitively conduct  complex multi-modal calculations, resulting in weak efficiency. Besides, ignoring the semantic relationship between different video-query pairs might miss important spatio-temporal semantic consistency (\textit{e.g.}, common noun/appearance ``woman'' and temporal relationship ``continues to'' in Figure \ref{fig:pipeline}), which limits their effectiveness.
Thus, an effective and efficient model is expected to explore the latent semantics relationship between different video-query pairs.

Hence, we pose a novel task: \emph{can we co-train multiple video-query pairs and transfer the grounding knowledge from a  pair to another  pair?} 
We demonstrate this brand-new task \emph{``multi-pair TSG''} (MP-TSG) in Figure~\ref{fig:intro}(c). 
To the best of our knowledge, there is no such setting proposed in existing works.
To address this brand-new and challenging setting, we propose a novel multi-thread framework to co-train different pairs. We notice that the semantic relationship between different video-query pairs includes four aspects: query-to-query relationship, video-to-query relationship, cross-modal spatial relationship (object-to-phrase) and cross-modal temporal relationship (activity-to-sentence).
Specifically, 
we first mine the  shared spatial  semantics and temporal relationships across different sentence queries to assist with each other in the TSG task. To mine the intra-modal information and obtain inter-modal representation simultaneously, we then design a cross-modal contrast module to explore the global-level semantic consistency between videos and queries by a self-supervised strategy. Moreover, we design an adaptive negative selection module to adaptively generate a dynamic threshold for cross-modal matching. To sufficiently align fine-grained visual information and fine-grained textual  information from spatial and temporal perspectives, we design a prototype alignment strategy to 1) match the object prototypes and phrase prototypes to align appearance representations across modalities, and 2) align activity prototypes and sentence  prototypes to integrate motion  representations between different modalities.  
Our main contributions are  as follows:
\begin{itemize}
    \item We pose and address a brand-new task: MP-TSG, which aims to co-train multiple video-query pairs by exploring the semantic relationships between 
    different pairs to assist with each other. We propose a novel multi-thread framework to co-train different pairs by mining the relationships in four aspects: query-to-query,  video-to-query, object-to-phrase, activity-to-sentence.
    \item We propose a novel cross-modal prototype alignment module to explore the semantic relationship between different queries/videos. Besides, to deeply explore  cross-modal matching, we design an adaptive negative selection module to automatically generate a dynamic threshold for semantically matching video-query pairs. 
    \item  Extensive experiments  on three challenging benchmarks 
    demonstrate both  effectiveness and efficiency of our  method. 
    More importantly, our method can serve  as a plug-and-play module for state-of-the-art methods to enhance their effectiveness and efficiency.
\end{itemize}

\section{Related Works}
\label{sec:related}

\noindent \textbf{Fully-supervised TSG.} 
Temporal sentence grounding (TSG) \cite{anne2017localizing,gao2017tall,ju2024deep,ju2020learning,ju2023gr,liang2024survey,liang2024simple,liang2023knowledge} aims at locating the most relevant segments from long videos corresponding to the given sentence descriptions. 
Traditional TSG methods \cite{gao2017tall,xiong2024rethinking,rao2021p0135,rao2021sm,jiang2023lttpoint,zhang2023deep,wang2025destination,wang2025taylor,tang2024reparameterization,tang2022decision,tang2021codes,tang2022rethinking,tang2023deep,tang2024cores,tang2024symattack} typically utilize a \textit{propose-and-rank} approach to make predictions based on interacted multi-modal features. 
Some \textit{proposal-free} methods \cite{zhang2020span,lin2024certifiably,lin2024decoding,lin2024trojan,lin2024stealing,wang2021hierarchical,wang2021pwclo,wang2020unsupervised} are proposed to directly predict the temporal locations of the target segment without generating proposals. 

\noindent \textbf{Weakly-supervised TSG.}
The above fully-supervised methods heavily rely on the datasets that require numerous manually labeled annotations for training.
To ease  human labeling efforts,  recent works \cite{mithun2019weakly,liu2023information,liu2022localized,liu2024masked,wen2023deep} consider a weakly-supervised setting to only access the information of matched video-query pairs without accurate segment boundaries.
However, their performance is less satisfactory with such weak supervision.

Many semantic relationships between different queries/videos are not explored in previous methods, leading to repeated training  and much computational cost. Unlike them, we introduce a brand-new setting,   MP-TSG, where different queries and videos can be co-trained to reduce the cross-modal gap between video and query. 
\section{Methodology}
\begin{figure*}[t!]
    \centering
    \includegraphics[width=0.96\textwidth]{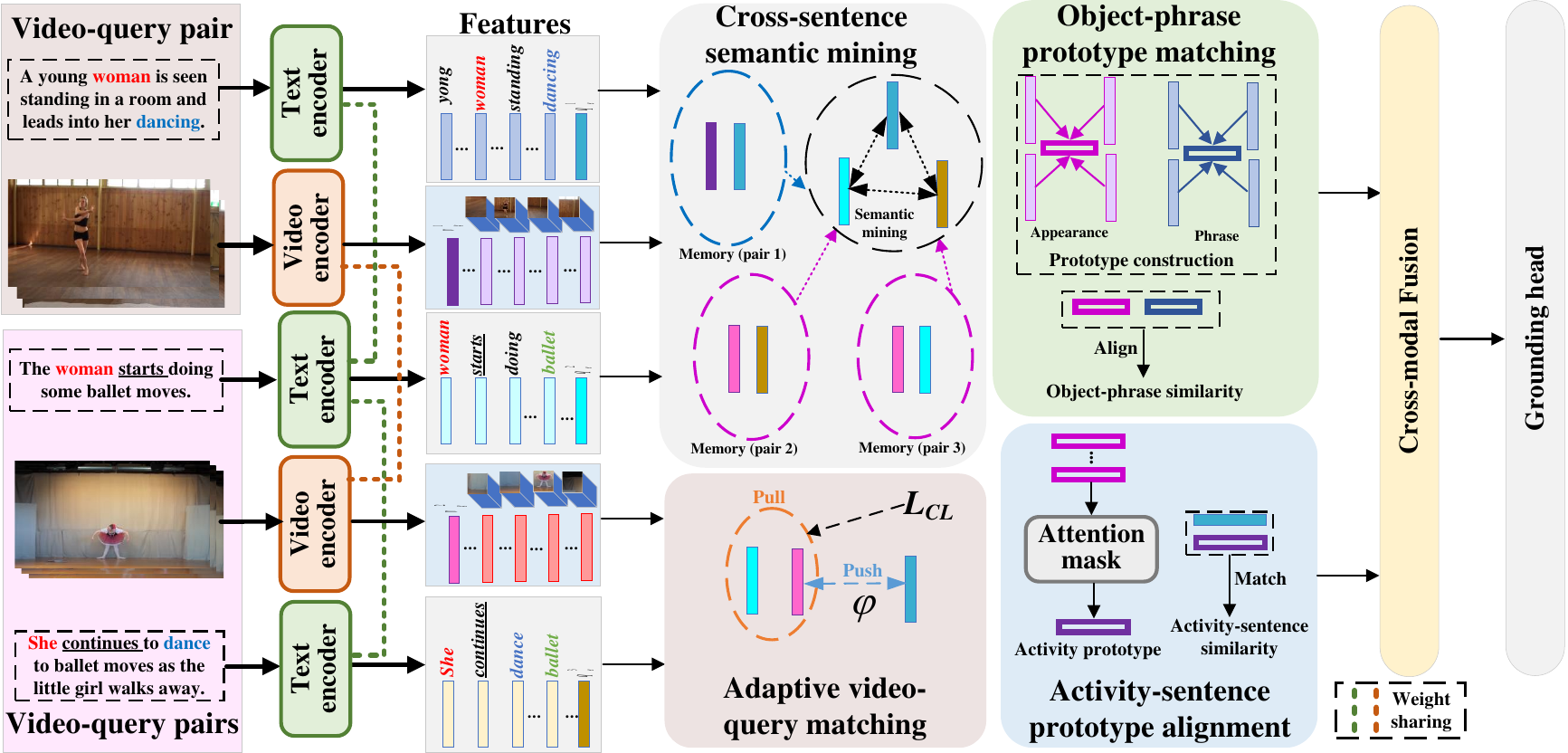}
\vspace{-9.5pt}
    \caption{\small Overview of our proposed MKTN for the  MP-TSG task.   Given some video-query pairs (\textit{e.g.,} the first and second videos correspond to one and two queries respectively), we first utilize video and query encoders to  extract corresponding features. Then, we feed these features into four carefully-designed modules to fully explore the semantic relationships between videos and queries. In the cross-sentence semantic mining module, we mine the query-to-query relationship based on the cross-modal memory. For the adaptive video-query matching module, we adaptively learn the cross-modal semantic consistency with video-to-query relationship
    by a dynamic threshold $\phi$ and a contrastive loss $\mathcal{L}_{CL}$. In the object-phrase prototype matching module, we align appearance representations across modalities based on appearance and phrase prototypes. Similarly, we  integrate motion  representations by aligning activity  and sentence  prototypes. 
    Best viewed in color.}
\vspace{-9.5pt}
    \label{fig:pipeline}
\end{figure*}

\noindent \textbf{Problem definition.}
Given $M_q$ video-query pairs $\{V_p,Q_p\}_{p=1}^{M_q}$, previous TSG methods aim to localize the query-described activity segment in the video for each video-query pair, where $V_p$ and $Q_p$ are the corresponding video and query, respectively. They independently regard each video-query pair with ignoring the semantic relationship between different queries and videos, and repeat the grounding process $M_q$ times.
Different from them, we pose a practical yet challenging setting,  Multi-Pair TSG (MP-TSG), which aims to co-train multiple video-query pairs for effective and efficient grounding.

\noindent \textbf{Pipeline.}
To tackle the MP-TSG task, we propose a novel framework in Figure~\ref{fig:pipeline}. The semantic relationship between different video-query pairs includes four aspects: query-to-query relationship, video-to-query relationship, cross-modal spatial relationship (object-to-phrase) and cross-modal temporal relationship (activity-to-sentence). The first two relationships are global-level, and we can determine whether  any video and any query are related or not. The last two relationships are local-level, which aligns the cross-modal semantics from spatial and temporal perspectives respectively for precise video grounding.

\subsection{Preparation}
\label{sec:encoder}

\noindent \textbf{Video encoder.}
Given $M_v$ videos $\{V_1, \cdots,V_{M_v}\}$, we first follow previous work \cite{gao2017tall} to extract its frame-wise features by a pre-trained 3D-CNN network \cite{tran2015learning}, and then employ a multi-head self-attention \cite{vaswani2017attention} module to capture the long-range dependencies among video frames. For the $a$-th video with $N_v$ frames, we denote the extracted video features as ${V}_a=\{{v}_g^i, {v}_{a_1}^i, \cdots, {v}_{a_C}^i\}_{i=1}^{N_v} \in \mathbb{R}^{N_v \times (C+1) \times d}$, where  $d$ is the feature dimension, $C$ is the patch number, ${v}_g^i$ is the global  feature of the $i$-th frame.

\noindent \textbf{Text encoder.}
Similarly, given $M_q$ queries $\{Q_1,\dots,Q_{M_q}\}$, 
by feeding any query $Q_j$ to the pretained Glove network \cite{pennington2014glove},
we can obtain the word-level features ${Q}_j=\{{q}_1^j,\cdots,{q}_{N_q}^j\} \in \mathbb{R}^{N_q \times d}$, where $N_q$ is the word number. To extract the semantic of the whole sentence, the Skip-thought parser~\cite{kiros2015skip} is employed to capture the  query-level feature ${q}_e^j\in\mathbb{R}^{d}$.
\subsection{Cross-Sentence Semantic Mining}
\label{cross-sentence}


Although previous TSG works \cite{anne2017localizing,gao2017tall} try to fully understand textual information and visual information \cite{zhao2024multi,xiong2024rethinking,cai2025imperceptible,liu2024towards,tang2025simplification,tang2024reparameterization,lei2025exploring,yang2025eood,liu2024pandora,zhang2025manipulating,zhang2025monoattack,hu2024exploiting,hu2025adaptive}, they often ignore the semantic relationship between different sentences.
The semantic relationship includes 1) the  temporal information between different segments in the same video, and 2) the contextual information among different sentences.
To sufficiently mine these query-to-query relationships, we  extract query-level features $F_q=\{q_e^j\}_{j=1}^{M_q}$ from the multiple queries rather than learnable embeddings in previous  works. Thus, we aim to model the  query-level contexts, and explore the temporally and contextually related queries of each query. For example, in Figure \ref{fig:pipeline}, ``The woman starts doing some ballet moves.'' and ``She continues to dance to ballet moves as the little girl walks away.'' share the same semantics (``woman'' and ``ballet moves''), and contains the temporal relationship (``start'' and ``continues to''). The shared semantics and temporal relationship will assist the grounding task of each sentence.

Given the query-level features $F_q$, we first encode temporal information by the position embedding layer, and then  conduct  the interactions among queries by the self-attention layers.  After that, we extract a textual feature for each  query to represent corresponding  events from the multi-modal memory by cross-attention layers. Considering that some sentences might share the same particular words (\textit{e.g.}, ``woman'' in Figure \ref{fig:pipeline}), we extract the hierarchical textual features to conduct the cross-granularity interactions, which makes the decoder learn more contextual information. Finally, we compute the timestamps of each query-wise feature by a parallel regression layer. The above procedures are formulated as 
$T=MLP(\tau(F_q,F_{mem}))$,
where $T=\{(t_s^j,t_e^j)\}_{j=1}^{M_q}$ denotes the ground-truth start and end timestamps ($t_s^j,t_e^j$) for $M_q$ queries;
$\tau$ denotes the decoder of the transformer to  conduct the query-level position embedding; $F_{mem}=[F_v^j;F_w^j]_{j=1}^{M_q}$ denotes the multi-modal memory, where $F_v^j=\{{v}_{a_1}^j, \cdots, {v}_{a_C}^j\}$ is the frame-level feature of the video paired with query $Q_j$,
$F_w^j=\{{q}_1^j, \cdots,{q}_{N_q}^j\}$ is the word-level feature, $[\cdot ; \cdot]$  denotes the concatenating operation.

\subsection{Adaptive Video-Query Matching}
\noindent \textbf{Adaptive negative selection.}
Common TSG datasets inherently treat a set of video-query pairs as positive matches. Negatives are assigned under the assumption that all non-corresponding pairs are semantically distinct. However, some queries labeled as negative may indeed partially or accurately align with a video, constituting false negatives. 
Therefore, we design a dynamic threshold-based negative selection strategy to adaptively select negatives. For a given query $Q_j$, the negative videos are selected as: 
\small
\begin{equation}\label{similar_v_q}
\mathcal{N}_q = \{V_i|S_{ij}<\phi\}\cap\mathcal{N}, {S}_{ij} = {V_i}({Q_j}{W}_S)^{\top} \in \mathbb{R}^{N_v \times N_q},
\end{equation}
\normalsize
where  $S_{ij}$  denotes the similarity between  video $i$ and query $j$; ${W}_S \in \mathbb{R}^{d \times d}$ projects the query features into the same latent space as the video; $\mathcal{N}$ is the original negative set; $\phi$ is a  dynamic threshold:
\small
\begin{equation}\label{adaptive_threshold}
\phi=\phi_{final}-(\phi_{final}-\phi_{intial})  \cos(r\pi+1),
\end{equation}
\normalsize
where $\phi_{intial}$ and $\phi_{final}$ are the thresholds at the start and end of training respectively,  $r\in [0,1]$ is the percentage of the training process. The design of the cosine annealing threshold where $\phi_{final}>\phi_{intial}$ is based on the intuition that the model has a higher confidence level in later training. In the early stages of training,  negatives with relatively high similarity scores can be reliably regarded as false negatives.

\noindent\textbf{Self-weighted cross-modal  contrast.}
\label{self-supervision}
To mine the intra-modal information and obtain inter-modal representations simultaneously, we  design a cross-modal contrast module to explore semantic consistency. As shown in Figure \ref{fig:pipeline}, we map the  word-level textual  feature $F_w$ and frame-level feature $F_v$ into a shared  subspace for semantic alignment. Especially, we introduce the transformer encoder $\varsigma$ to generate the transferred word-level textual  feature $F'_w$ and video feature $F'_v$ as: $F'_w=Norm(\varsigma(F_w)), F'_v=Norm(\varsigma(F_v))$.

We  construct a triplet tuple $(F'^+_v,F'^+_w,F'^-_w)$ to denote the pair relationship across queries, where $(F'^+_v,F'^+_w)$  is a positive pair and $(F'^+_v,F'^-_w)$ is a  negative pair.
 To pull the positive pairs $(F'^+_v,F'^+_w)$ together and push the negative pairs $(F'^+_v,F'^-_w)$ away, we map the  word-level textual  feature $F_w$ and frame-level feature $F_v$ into a shared  subspace for semantic alignment. Similarly, we can obtain the inter-video relationship: $(F'^+_w,F'^+_v,F'^-_v)$, where $(F'^+_w,F'^+_v)$  is a positive pair and $(F'^+_w,F'^-_v)$ is a  negative pair.
 Thus, we design the following contrastive loss for self-supervision:
 \small
\begin{align}\label{L_CL}
&\mathcal{L}_{CL}\!=\!\sum\nolimits_{F'^+_v,F'^+_w}\!\{\theta\!\sum\nolimits_{F'^-_w}\!\max[0,\!\phi\!-\!S_{(F'^+_v,F'^+_w)}\!+\!S_{(F'^+_v,F'^-_w)}]\nonumber\\
&+(1-\theta)\sum\nolimits_{F'^-_v}\max[0,\phi-S_{(F'^+_w,F'^+_v)}+S_{(F'^+_w,F'^-_v)}]\},
\end{align}
\normalsize
 where $\phi$ is the dynamic threshold in Eq. \eqref{adaptive_threshold}; the scoring function $S_{(\cdot,\cdot)}$ measures the similarity between the visual feature and textual feature in the joint space; $\theta\in(0,1)$ is a regularized parameter to balance the significance between negative videos and negative queries. We can obtain $\theta$ by:
  \small
\begin{align}\label{mu}
\frac{\theta}{1-\theta}=\frac{(||F'^+_v-F'^-_w||_2-||F'^+_v-F'^+_w||_2)^2}{(||F'^+_v-F'^-_w||_2-||F'^+_v-F'^+_w||_2)^2}.
\end{align}
\normalsize

\subsection{Object-Phrase Prototype Matching}
\label{objeect_phrase}

The query subjectively describes a video activity, resulting in partially matched video-query pairs. Although a single patch-activity projection is intuitive for query-adaptive matching, it falls short in capturing local details. Decoupling the spatio-temporal modeling process in a divide-and-conquer manner proves more effective. Hence, we design   progressive object-phrase  prototype matching.


\noindent \textbf{Constructing appearance prototype.} For the given videos, we aggregate the patch features into object-level prototypes to represent fine-grained appearance representations, such as object instance, object part, and action region. During prototype construction, not all patch features are aggregated. Although the patch features contain important appearance representations, they also bring redundancy. For example, some background regions  may interfere with cross-modal alignment. Hence, we filter out retrieval-superfluous information and generate object-level prototypes in a sparse aggregation manner. For convenience, three Fully Connected (FC) layers and ReLU function are utilized to predict sparse visual weights $W_a^i\in \mathbb{R}^{(C+1)\times N_a}$, where $N_a$ denotes the number of object-level prototypes in the $i$-th video. Therefore, we prevent these object-level prototypes from being affected by redundant patches. 

For each frame $F_v^i\in \mathbb{R}^{(C+1)\times d}$, the constructed appearance prototype is defined as $P_a^i=W_a^{i^T} F_v^i\in \mathbb{R}^{N_a\times d}$. Ideally, each object prototype can adaptively aggregate the corresponding object-related or action-related patches.
As for each phrase, we also apply a similar network structure to aggregate word features. Similarly, we utilize three FC layers and ReLU functions to obtain sparse textual weights $W_p^j\in \mathbb{R}^{(C+1)\times N_p}$, where $N_p$ is the number of phrase prototypes in the $j$-th query. Thus, we can fully extract the significant appearance representations by the fine-grained patch features and word features. Besides, the phrase prototypes $P_p^{i,j}=W_p^{j^T}F_v^i$ are optimized by spatial object-phrase prototype matching.

\noindent \textbf{Object-phrase cross-modal alignment.}  We propose a prototype-wise query-video interaction from an appearance perspective.  Specifically, we first compute the maximum similarity of object-phrase prototypes within each frame. This associates the phrase prototypes most similar to each object prototype, reflecting cross-modal fine-grained alignment. Then, for the multi-frame object similarity matrix, we find the largest similarity score across the frame sequences, which gives a more confident probability of object-phrase matching. Finally, the object-phrase matching scores are summed for the final similarity: $s_a=\frac{1}{N_a}\sum_{l=1}^{N_a}\max_{i=1}^{N_v}\max_{j=1}^{N_p}[P_p^{i,j} \times P_a^{l^T}]$.

\subsection{Activity-Sentence Prototype Alignment}
\label{activity_sentence}
To  model the motion information, we explore the activity-to-sentence relationship: 1) first perform progressive activity-sentence prototype aggregation to reveal the video semantic diversity, 2) then design dynamic prototype alignment.

\noindent \textbf{Building activity prototype.} To fully understand the video activity, we aim to design  diverse activity prototypes.
A naive solution to obtain video-level features based on global frame features is by mean pooling, or by adding motion encoder layers. However, this leads to two issues: 1) failure to perceive local details and ignoring important objects will exacerbate the bias of video feature learning; 2) these strategies generate a single video-level feature, which can only quantify one-to-one relations. Therefore, we investigate how to incorporate key fine-grained objects and dynamic motion changes into diverse activity prototypes.

The core idea is to progressively aggregate spatial object prototypes into frame prototypes and then perform inter-frame interaction to generate various activity prototypes. A frame decoder is first designed to incorporate all object prototypes $P_a\in \mathbb{R}^{(N_v\times N_a)\times d}$ into frame-level prototypes $P_f\in \mathbb{R}^{N_a\times d}$, which implies fine-grained inter-object spatial relations. To learn frame-level object relations, we define the masked attention as: $P_f=Q_f+softmax(Q_fC_a^T+W_f)V_a$, where $Q_f\in \mathbb{R}^{N_a\times d}$ refers to frame queries (\textit{i.e.}, a set of randomly initialized learnable features), $V_a$ and $C_a^T$ are the features after the linear transformation of object prototypes $P_a$. The attention mask $W_f\in \mathbb{R}^{N_v\times(N_v\times N_a)}$ is: 
\small
\begin{align}
W_f(i,j) =\left\{\begin{matrix}0&\text{ if } N_a\cdot i\leq J<N_a\cdot (i+1),\\ -\infty&\text{otherwise.} \end{matrix}\right.
\end{align}
\normalsize
We add frame prototype $p_f^i$ original global feature $v_C^i$ of corresponding frames to enhance the robustness of the model: $p_f^i=(p_f^i+v_C^i)/{2}$.

Next, a dynamic activity decoder is developed to learn the inter-frame relationship in $P_f$,  which can obtain different activity prototypes $P_e\in \mathbb{R}^{M_q\times d}$ to illustrate the rich information of videos. Our dynamic attention is formulated as: $P_e=Q_e+softmax(Q_eK_a^T)V_a$,
where $Q_e=[q_e^1,q_e^2,\cdots,q_e^{M_q}]\in \mathbb{R}^{M_q\times d}$ refers to activity queries, $V_f$ and $K_f^T$ are the features after the linear transformation of object prototypes $P_f$. During training, each activity query learns how to adaptively focus on video frame prototypes, while multiple queries implicitly guarantee a certain activity diversity.
Differently, since the same video often corresponds to multiple text semantic descriptions, we directly use the global text representation $q_e$ as a sentence prototype to align with the activity prototypes  $P_e$.

\noindent \textbf{Activity-sentence alignment.} The activity-sentence prototype alignment is expressed as: $s_{es}\!=\!\max_{i=1}^{M_q}(q_e,P_{e_i})$.
By the similarity, we can find the closest activity prototype to the text representation for dynamic alignment.

\subsection{Multi-Modal Fusion and Grounding Head}
\label{cross_modal_fusion}

After obtaining the video feature ${V}_i$ and query feature ${Q}_j$, we further utilize a co-attention mechanism \cite{lu-etal-2019-debug} to capture the cross-modal interactions between videos and queries. Specifically, we first calculate the similarity $S_{ij}$ between ${V}_i$ and ${Q}_j$ by Eq. \eqref{similar_v_q}.
Then, we compute two attention weights as: ${A} = {S}_r ({Q}{W}_S) \in \mathbb{R}^{N_v \times d}$ and ${B} = {S}_r {S}_c^{\text{T}} {V} \in \mathbb{R}^{N_v \times d}$,
where ${S}_r$ and ${S}_c$ are the row- and column-wise softmax results of ${S}$, respectively. We compose the final query-guided video representation by learning its sequential features: ${F} = \text{Bi-GRU}([{V};{A};{V}\odot {A};{V}\odot {B}]) \in \mathbb{R}^{N_v \times d}$,
where $\text{Bi-GRU}(\cdot)$ denotes the Bi-GRU layers, 
and $\odot$ is the element-wise multiplication. The output ${F} = \{{f}^{i}\}_{i=1}^{N_v}$ encodes visual features with query-guided attention, where ${f}^i \in \mathbb{R}^{d}$.

In our model, we treat query-video pairs as positive examples, while considering all other pairwise combinations in the batch as negative examples. To fully leverage the query-video pair information, we introduce the query-to-video Robust InfoNCE (RINCE) loss as follows: 
\small
\begin{equation}
\mathcal{L}_{q\rightarrow v}(S)\!=\!\frac{1}{M_v}\!\sum\nolimits_{i=1}^{M_v}[-\frac{\exp(S^{ii})}{\tau}\!+\!\frac{(\alpha\sum\nolimits_{j=1}^{M_q}\exp(S^{ij}))^\tau}{\tau}\!],
\end{equation}
\normalsize
where $\tau,\alpha \in (0,1]$  are learnable  parameters. Similarly, the video-to-query loss is:
\small
\begin{equation}
\mathcal{L}_{v\rightarrow q}(S)\!=\!\frac{1}{M_q}\!\sum\nolimits_{j=1}^{M_q}[\!-\!\frac{\exp(S^{jj})}{\tau}\!+\!\frac{(\alpha\sum\nolimits_{i=1}^{M_v}\exp(S^{ji}))^\tau}{\tau}].
\end{equation}
\normalsize
Denoting activity-sentence and object-phrase prototype matching similarity matrices as $S_{es}$ and  $S_{op}$ respectively, we design the alignment loss as follows:
\small
\begin{equation}
\mathcal{L}_1=\mathcal{L}_{v\rightarrow q}(S_{es})+\mathcal{L}_{v\rightarrow q}(S_{op})+\mathcal{L}_{q\rightarrow v}(S_{es})+\mathcal{L}_{q\rightarrow v}(S_{op}).
\end{equation}
\normalsize

To predict the segment start/end boundary quickly and accurately, we first introduce the  span predictor \cite{zhang2020span} with two stacked transformer blocks and two feed-forward layers. Specifically, the multi-modal feature ${F}$ is fed into the span predictor, followed by a softmax function, to obtain two probability scores of start and end boundaries. We denote them as ${P}_{s(e)} \in \mathbb{R}^{N_v}$. The rounded integer boundaries $\hat{t}_{s(e)}$ are used to generate one-hot label vectors ${Y}_{s(e)}$ as  supervision.
\small
\begin{equation}\label{eq13}
    \mathcal{L}_{2} = L_{CE}({P}_{s}, {Y}_{s})+L_{CE}({P}_{e}, {Y}_{e}),
\end{equation}
\normalsize
where $L_{CE}$ means the cross-entropy loss.
The predicted boundary timestamps $\hat{t}_{s(e)}'$ are obtained by $P_{s(e)}$: $(\hat{t_{s}}', \hat{t_{e}}') = arg \max_{\hat{t}_{s}', \hat{t}_{e}'} {P}_{s}(\hat{t_{s}}') {P}_{e}(\hat{t}_{e}')$,
where $0 \le \hat{t}_{s}' \le \hat{t}_{e}' \le N_v$.

Since the above span predictor can only predict coarse integer boundary values, 
we additionally design a parallel float predictor consisting of several feed-forward layers to provide fine-grained float boundaries by the following loss: 
\small
\begin{equation}\label{eq15}
    \mathcal{L}_3 = f_{\text{L1-smooth}}(\hat{t}_{s}', t_s)+f_{\text{L1-smooth}}(\hat{t}_{e}',t_e),
\end{equation}
\normalsize
where $f_{\text{L1-smooth}}$ represents the smooth L1 loss and $(t_s,t_e)$ is the ground-truth boundary.
The predicted float boundaries ${O}_{s}$ and ${O}_{e}$ respectively represent the percentage of start and end boundary frames that are query-relevant.
Therefore, the fine-grained boundary indexes $\tilde{t}_{s(e)}'$ are calculated by: $(\tilde{t}_s', \tilde{t}_e') = (\hat{t}_{s}'+1-{O}_s,\hat{t}_{e}'-1+{O}_e)$.

The multi-modal network is trained  by minimizing the weighted sum of the above losses, denoted as $\mathcal{L}$:
\small
\begin{equation}\label{eq27}
    \mathcal{L} = \mathcal{L}_{CL}+\lambda\mathcal{L}_{1} + \gamma \mathcal{L}_{2}+\mu \mathcal{L}_{3},
\end{equation}
\normalsize
where $\lambda,\gamma$ and $\mu$ are parameters to weigh different losses.

\section{Experiments}

\begin{table}[t!]
\small
    \centering
    \setlength{\tabcolsep}{2mm}{
    \begin{tabular}{c|c|cccc}
    \hline
   \multicolumn{6}{c}{ActivityNet Captions}\\ \hline
    \multirow{2}*{Method} & \multirow{2}*{Type} & R@1, & R@1, & R@5, & R@5, \\ 
    ~ & ~ & IoU=0.5 & IoU=0.7 & IoU=0.5 & IoU=0.7\\ \hline 
    MRTNet&FS&42.02& 24.25&-&-\\
    2D-TAN  & FS & 44.51 & 26.54 & 77.13 & 61.96 \\
    MMN &FS&48.59& 29.26&79.50& 64.76\\ 
    VDI& FS&48.09& 28.76& 79.69& 64.88\\
\hline
ICVC  & WS &  29.52&- & 66.61&- \\
VCA & WS &  31.00&- &  53.83&- \\
WSTAN & WS &  30.01&- &  63.42&-\\
CPL & WS&31.37&-&  43.13&-\\
MMDist&WS&32.98&-&-&-\\
\hline
\textbf{Ours} & \textbf{MP} & \textbf{58.32}& \textbf{35.28}& \textbf{86.20} & \textbf{71.49} \\\hline
    \hline
   \multicolumn{6}{c}{Charades-STA}\\ \hline
     \multirow{2}*{Method} & \multirow{2}*{Type} & R@1, & R@1, & R@5, & R@5, \\ 
~ & ~ & IoU=0.5 & IoU=0.7 & IoU=0.5 & IoU=0.7\\
\hline
      VDI&FS&52.32& 31.37& 87.03& 62.30\\
 DRN  & FS & 53.09 & 31.75 & 89.06 & 60.05 \\
 MESM&FS& 56.69& 35.99&-&-\\
 MRTNet&FS&62.50& 43.63&-&-\\
\hline
ICVC  & WS & 31.02 & 16.53 & 77.53 & 41.91 \\
VCA & WS & 38.13 & 19.57 & 78.75 & 37.75 \\
LCNet  & WS & 39.19 & 18.17 & 80.56 & 45.24 \\
CPL&WS& 49.24& 22.39& 84.71 & 52.37\\
MMDist&WS&54.72& 26.00&-&-\\
    \hline 
    \textbf{Ours} & \textbf{MP}& \textbf{69.93}& \textbf{46.27}& \textbf{97.16}& \textbf{69.13} \\ \hline
    \hline
   \multicolumn{6}{c}{TACoS}\\ \hline
\multirow{2}*{Method} & \multirow{2}*{Type} & R@1, & R@1, & R@5, & R@5, \\ 
    ~ & ~ & IoU=0.3 & IoU=0.5 & IoU=0.3 & IoU=0.5  \\ \hline 
 DRN &FS&-&23.17&-&33.36\\
2D-TAN &FS&37.29&25.32&57.81&45.04\\
MRTNet & FS& 37.81& 26.01&-&-\\
MMN &FS&39.24&26.17&62.03&47.39\\
 MIGCN &FS&48.79& 37.57& 67.63& 57.91\\
 MESM&FS& 52.69& 39.52&-&-\\
    \hline 
    \textbf{Ours} & \textbf{MP}& \textbf{53.38}& \textbf{42.62}& \textbf{73.54}& \textbf{62.24}\\ \hline
 \end{tabular}}
      \vspace{-12pt}
      \caption{\small Effectiveness comparison for TSG on all the datasets under official train/test splits, where 
     ``MP'' means ``Multi-Pair''.}
\vspace{-8pt}
    \label{tab:main}
\end{table}

\noindent \textbf{Datasets.}
For a fair comparison with previous works \cite{zhang2019learning,wang2022negative,zheng2022weakly}, we utilize the same 
ActivityNet Captions \cite{caba2015activitynet}, TACoS \cite{regneri2013grounding}, and Charades-STA \cite{sigurdsson2016hollywood} datasets for evaluation. 

\noindent \textbf{Evaluation metrics.}
Following \cite{gao2017tall}, we evaluate the grounding performance by ``R@n, IoU=m'', which means the percentage of  queries having at least one result whose Intersection over Union (IoU) with ground truth is larger than m. 

\noindent \textbf{Compared methods.} 1) Fully-supervised (FS): DRN \cite{zeng2020dense}, 2D-TAN \cite{zhang2019learning},  MIGCN \cite{zhang2021multi}, MMN \cite{wang2022negative}, VDI \cite{luo2023towards},   MRTNet \cite{ji2024mrtnet}, MESM \cite{liu2024towards}.
2) Weakly-supervised (WS):  LCNet  \cite{yang2021local}, VCA \cite{wang2021visual}, ICVC  \cite{chen2022explore}, WSTAN \cite{wang2022weakly},  CNM \cite{zhengming2022weakly}, CPL \cite{zheng2022weakly}, MMDist \cite{bao2024local}.

\begin{table}[t!]
    \small
    \centering
        \setlength{\tabcolsep}{1mm}{
    \begin{tabular}{c|c|cccc|ccc}
    \toprule
    \multirow{2}*{Methods} & \multirow{2}*{Variant} & R@1, & R@1, & R@5, & R@5, & \multirow{2}*{VPS}\\ 
    ~ & ~ & IoU=0.5 & IoU=0.7& IoU=0.5 & IoU=0.7 \\  \midrule
    \multirow{2}*{2D-TAN } & Origin & 39.81 & 23.25& 79.33& 52.15 &12.43 \\
    ~ & \textbf{+Ours} & \textbf{43.37} & \textbf{27.48} & \textbf{86.74}& \textbf{55.68}& \textbf{46.28}\\ \midrule
    \multirow{2}*{MMN } & Origin & 47.31& 27.28&83.74& 58.41& 21.63  \\
    ~ & \textbf{+Ours} & \textbf{49.36} & \textbf{31.03} & \textbf{88.75}& \textbf{62.39}&\textbf{48.38}\\ \midrule
    \multirow{2}*{LCNet } & Origin & 39.19 & 18.17 &80.56& 45.24& 23.88 \\
    ~ & \textbf{+Ours} & \textbf{42.80} & \textbf{25.94} &\textbf{82.46}& \textbf{51.37}& \textbf{52.73}\\ \bottomrule
    \end{tabular}}
    \vspace{-10pt}
    \caption{\small We serve our method as a plug-and-play module for  state-of-the-art TSG methods on  Charades-STA under  official train/test splits, where ``VPS'' denotes ``video per second'' during inference.}
     \label{tab:plug_and_play}
\vspace{-10pt}
\end{table}

\begin{figure*}[t!]
\centering
\includegraphics[width=\textwidth]{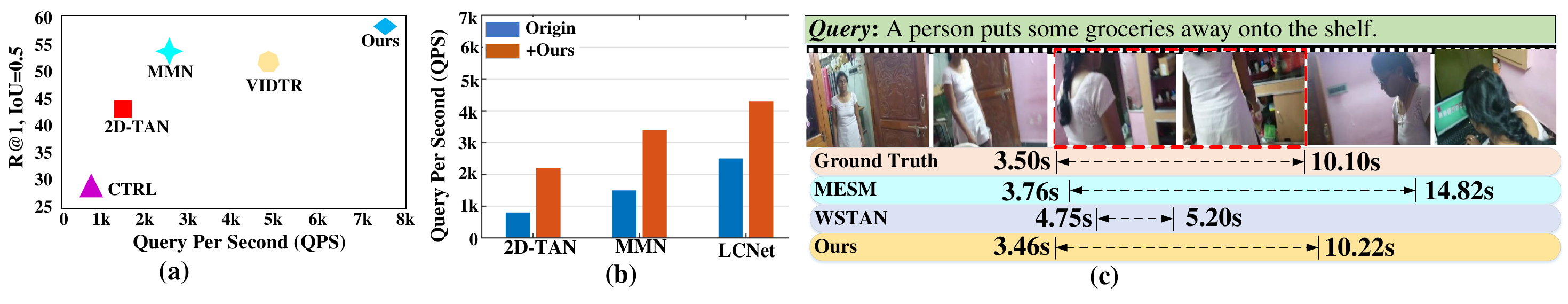}
\vspace{-20pt}
\caption{\small Performance comparison with state-of-the-art methods on Charades-STA. (a) compares the effectiveness (R@1, IoU=0.5) and the efficiency (QPS), (b) shows that our method can serve as a plug-and-play module to enhance their efficiency, (c) is the  qualitative results.}
\label{fig:vis_result}
\vspace{-5pt}
\end{figure*}

\begin{table*}[t!]
\small
\scalebox{1.0}{
\setlength{\tabcolsep}{1.3mm}{
\begin{tabular}{c|cccc|cccc|cccccccccccc}
\hline
\multirow{3}*{Model}&\multicolumn{4}{|c|}{ActivityNet Captions} & \multicolumn{4}{|c|}{Charades-STA} &\multicolumn{4}{|c}{TACoS}\\\cline{2-13}
& R@1 & R@1 & R@5 & R@5& R@1 & R@1 & R@5 & R@5& R@1 & R@1 & R@5 & R@5\\
 & IoU=0.5 & IoU=0.7 & IoU=0.5 & IoU=0.7& IoU=0.5 & IoU=0.7 & IoU=0.5 & IoU=0.7& IoU=0.3 & IoU=0.5 & IoU=0.3 & IoU=0.5\\\hline
w/o CSM& 53.82& 28.60& 81.95& 67.76& 65.82& 44.45& 92.11& 63.19& 47.38& 33.12& 68.44& 56.13\\
w/o OPM & 55.40& 30.33& 83.95&70.02&59.80& 39.13& 93.32& 62.01& 49.85& 37.24& 69.48& 59.16\\
w/o APA& 56.27& 30.86& 80.21& 68.49& 60.77& 42.96& 90.11& 65.24& 50.92& 37.25& 70.42& 59.08 \\
w/o AVM &54.95 &31.03 &83.72&  69.30& 65.21& 45.75& 93.24& 67.69& 50.46& 38.05& 70.85& 58.49 \\\hline
\textbf{Full} &\textbf{58.32}& \textbf{35.28}& \textbf{86.20} & \textbf{71.49} & \textbf{69.93}& \textbf{46.27}& \textbf{97.16}& \textbf{69.13}&\textbf{53.38}& \textbf{42.62}& \textbf{73.54}& \textbf{62.24}\\ \hline
\end{tabular}}}
\vspace{-10pt}
\caption{\small Main ablation study, where we remove each key individual component to investigate its effectiveness. ``CSM'' denotes ``cross-sentence semantic mining'',  ``OPM'' denotes ``object-phrase prototype matching'', ``APA'' denotes ``activity-sentence prototype alignment'', ``AVM'' denotes ``adaptive video-query matching'', ``Full'' denotes our full model.}
\vspace{-10pt}
\label{tab:ablation1}
\end{table*}


\subsection{Comparison With State-Of-The-Arts}

\noindent \textbf{Quantitative comparison.}
We compare our proposed method with other existing state-of-the-art approaches  in Table \ref{tab:main} and Figure \ref{fig:vis_result}(a). 
Obviously, our proposed method outperforms both fully-supervised and weakly-supervised methods by a large margin. The main reasons are as follows: 
1) ActivityNet Captions: these sentence annotations share many nouns and sequencing words, and our method can mine the spatial and temporal relationship between different sentences by our cross-sentence semantic mining module. 
2) Charades-STA: some video-query pairs share the same sentence queries. Thus, our method can co-train these pairs and transfer the knowledge from some easy pairs to difficult pairs. 
3) TACoS: TACoS only contains activities of cooking scenarios, where these videos often share  similar object/appearance information. Our method leverages the information to co-train different videos for better video understanding.

\noindent \textbf{Plug-and-play.} 
Besides, our method can serve as a plug-and-play module for state-of-the-art models (fully-supervised TSG: 2D-TAN and MMN, weakly-supervised TSG: LCNet).
As shown in Table \ref{tab:plug_and_play} and Figure \ref{fig:vis_result}(b), our method can significantly improve their performance with higher efficiency  on the Charades-STA dataset.
Therefore, our  multi-thread framework for TSG is flexible and can be adopted into other state-of-the-art methods to improve their effectiveness and efficiency. The performance improvement is because 1) effectiveness: our multi-thread framework can mine the semantic relationship between different queries, which allows them to assist each other for grounding. 2) efficiency: our framework can co-train multiple video-query pairs to reduce redundant calculations. 

\noindent \textbf{Visualization comparison.}
To qualitatively investigate the effectiveness of our method, we report some representative examples in Figure \ref{fig:vis_result}(c), where  our grounding result is closer to the ground truth than MESM and WSTAN.

\subsection{Ablation Study}

\noindent \textbf{Main ablation studies.}
To evaluate the contribution of each module, we perform the main ablation study in Table~\ref{tab:ablation1}.  All the modules contribute a lot to the final performances, demonstrating  their effectiveness in exploring the intra- and inter-modal relationship in multiple video-query pairs.

\begin{table}[t!]
\small
\scalebox{1.0}{
\setlength{\tabcolsep}{0.6mm}{
\begin{tabular}{c|cccccccccccccccccccc}
\hline
\multirow{2}*{Model}& R@1 & R@1 & R@5 & R@5\\
& IoU=0.5 & IoU=0.7 & IoU=0.5 & IoU=0.7\\\hline
w/o Temporal relationship&56.40& 30.98& 78.36& 69.88 \\
w/o Contextual relationship& 57.14& 31.06& 80.12& 67.84\\
w/o Semantic relationship & 54.85&30.81& 78.48& 68.95
\\ \hline
\textbf{Full} &\textbf{58.32}& \textbf{35.28}& \textbf{86.20} & \textbf{71.49} \\\hline
\end{tabular}}}
\vspace{-10pt}
\caption{\small Ablation study on cross-sentence semantic mining.}
\vspace{-5pt}
\label{tab:ablation_csm}
\end{table}


\noindent \textbf{Importance of    cross-sentence semantic mining (CSM).} To assess the effectiveness  of our CSM module, we compare different ablation models on  ActivityNet Captions
in Table \ref{tab:ablation_csm}, where we remove one relationship between different  sentence queries in the first three ablation models. Obviously, our full model achieves the best performance because ActivityNet Captions contains a large number of  semantically related queries, and our model can fully mine the semantic relationship between different queries for grounding.


\begin{table}[t!]
\small
\scalebox{1.0}{
\setlength{\tabcolsep}{1.5mm}{
\begin{tabular}{cc|ccccccccccccccccccc}
\hline
\multicolumn{6}{c}{Cross-modal contrast via self-supervision}\\\hline
Visual&  Textual& R@1 & R@1 & R@5 & R@5\\
feature& feature& IoU=0.5 & IoU=0.7 & IoU=0.5 & IoU=0.7\\\hline
\XSolidBrush &\CheckmarkBold & 53.03& 29.96& 80.31& 68.69\\
\CheckmarkBold & \XSolidBrush &52.84& 30.12& 79.68& 67.45 \\\hline
\CheckmarkBold &\CheckmarkBold  &\textbf{58.32}& \textbf{35.28}& \textbf{86.20} & \textbf{71.49}\\ \hline\hline
\multicolumn{6}{c}{Adaptive negative selection}\\\hline
\multicolumn{2}{c|}{ Fixed threshold}& 54.08& 31.09& 82.20& 65.13 \\
\multicolumn{2}{c|}{w/o $\cos(p\pi+1)$}& 55.15& 31.26& 82.39& 68.04 \\\hline
\multicolumn{2}{c|}{\textbf{w/ $\cos(p\pi+1)$}}&\textbf{58.32}& \textbf{35.28}& \textbf{86.20} & \textbf{71.49} \\\hline
\end{tabular}}}
\vspace{-8pt}
\caption{\small Ablation study on adaptive video-query matching.}
\vspace{-10pt}
\label{tab:ablation_cce}
\end{table}

\noindent \textbf{Influence of cross-modal contrast (CC).} To analyze the importance of our CC module, we conduct corresponding experiments on  ActivityNet Captions  in Table \ref{tab:ablation_cce}. Both visual and textual features contribute a lot to integrating different video-query pairs. It is because our model can use the self-supervision strategy in Eq. \eqref{L_CL} to fully  mine the intra- and inter-modal representations for video grounding.

\noindent \textbf{Effect of  adaptive negative selection (ANS).} To assess the performance of our ANS module, we change the threshold to obtain two ablation models in Table \ref{tab:ablation_cce}. 
Obviously, our full model obtains the best results since our ANS module can generate an adaptive threshold for negative query selection, which can fully match queries and relevant videos.

\begin{table}[t!]
\small
\scalebox{1.0}{
\setlength{\tabcolsep}{2.2mm}{
\begin{tabular}{cc|ccccccccccccccccccc}
\hline
\multirow{2}*{CAP}&  \multirow{2}*{OCA}& R@1 & R@1 & R@5 & R@5\\
~& ~& IoU=0.5 & IoU=0.7 & IoU=0.5 & IoU=0.7\\\hline
\XSolidBrush &\CheckmarkBold & 56.80& 31.03& 83.16& 70.93 \\
\CheckmarkBold & \XSolidBrush &57.03& 30.72& 84.13& 69.48 \\\hline
\CheckmarkBold &\CheckmarkBold  &\textbf{58.32}& \textbf{35.28}& \textbf{86.20} & \textbf{71.49}\\ \hline
\end{tabular}}}
\vspace{-8pt}
\caption{\small Ablation study on object-phrase prototype matching, where ``CAP'' denotes ``constructing appearance prototype'' and ``OCA'' denotes ``object-phrase cross-modal alignment''.}
\vspace{-5pt}
\label{tab:ablation_opm}
\end{table}

\noindent \textbf{Significance of object-phrase prototype matching (OPM).} To analyze the performance of our OPM module in integrating the cross-modal spatial information, an ablation experiment is conducted on ActivityNet Captions  in Table \ref{tab:ablation_opm}. Our full model beats other ablation models by a large margin since  appearance and phrase prototypes provide visual and textual spatial semantics for multi-modal fusion.

\begin{table}[t!]
\small
\scalebox{1.0}{
\setlength{\tabcolsep}{2.4mm}{
\begin{tabular}{cc|ccccccccccccccccccc}
\hline
\multirow{2}*{BMP}&  \multirow{2}*{AA}& R@1 & R@1 & R@5 & R@5\\
~& ~& IoU=0.5 & IoU=0.7 & IoU=0.5 & IoU=0.7\\\hline
\XSolidBrush &\CheckmarkBold & 54.88& 30.91& 82.42& 67.60 \\
\CheckmarkBold & \XSolidBrush &53.04& 31.09& 83.86& 68.95 \\\hline
\CheckmarkBold &\CheckmarkBold  &\textbf{58.32}& \textbf{35.28}& \textbf{86.20} & \textbf{71.49}\\ \hline
\end{tabular}}}
\vspace{-10pt}
\caption{\small Ablation study on activity-sentence prototype alignment, where ``BMP'' denotes ``building motion prototype'' and ``AA'' denotes ``activity-sentence alignment''.}
\vspace{-10pt}
\label{tab:ablation_apa}
\end{table}

\noindent \textbf{Analysis on activity-sentence prototype alignment (APA).} We further analyze the  performance of our APA module for cross-modal temporal representations on  ActivityNet Captions  in Table \ref{tab:ablation_apa}. Obviously, both modules bring significant performance improvement since  activity and sentence prototypes can be used to understand the temporal semantics from visual and textual perspective respectively.
\section{Conclusion}\label{conc}
In this paper, we pose a brand-new and realistic setting: Multi-Pair TSG. For the challenging task, we propose a novel Multi-Thread Knowledge Transfer Network (MKTN) to deeply explore  intra- and inter-modal relationships.  
Extensive experiments on three challenging datasets show the effectiveness and efficiency of our  MKTN. 
Moreover, our MKTN can serve as a plug-and-play module for previous methods to enhance their effectiveness and efficiency.

{\small
\bibliography{cvpr24}

@article{liu2020saanet,
  title={SAANet: Siamese action-units attention network for improving dynamic facial expression recognition},
  author={Liu, Daizong and Ouyang, Xi and Xu, Shuangjie and Zhou, Pan and He, Kun and Wen, Shiping},
  journal={Neurocomputing},
  year={2020}
}

@article{fang2025your,
  title={Your data is not perfect: Towards cross-domain out-of-distribution detection in class-imbalanced data},
  author={Fang, Xiang and Easwaran, Arvind and Genest, Blaise and Suganthan, Ponnuthurai Nagaratnam},
  journal={Expert Systems with Applications},
  year={2025}
}

@article{fang2023hierarchical,
  title={Hierarchical local-global transformer for temporal sentence grounding},
  author={Fang, Xiang and Liu, Daizong and Zhou, Pan and Xu, Zichuan and Li, Ruixuan},
  journal={IEEE Transactions on Multimedia},
  year={2023},
  publisher={IEEE}
}

@article{fang2022multi,
  title={Multi-modal cross-domain alignment network for video moment retrieval},
  author={Fang, Xiang and Liu, Daizong and Zhou, Pan and Hu, Yuchong},
  journal={IEEE Transactions on Multimedia},
  volume={25},
  pages={7517--7532},
  year={2022},
  publisher={IEEE}
}

@inproceedings{fang2026cogniVerse,
  title={CogniVerse: Revolutionizing Multi-modal Retrieval-Augmented Generation with Cognitive Reflection and Geometric Reasoning},
  author={Fang, Xiang and Fang, Wanlong and Wang, Changshuo},
  booktitle={Proceedings of the IEEE/CVF Conference on Computer Vision and Pattern Recognition},
  year={2026}
}

@inproceedings{fang2023you,
  title={You can ground earlier than see: An effective and efficient pipeline for temporal sentence grounding in compressed videos},
  author={Fang, Xiang and Liu, Daizong and Zhou, Pan and Nan, Guoshun},
  booktitle={Proceedings of the IEEE/CVF Conference on Computer Vision and Pattern Recognition},
  pages={2448--2460},
  year={2023}
}

@inproceedings{fang2025hierarchical,
  title={Hierarchical Semantic-Augmented Navigation: Optimal Transport and Graph-Driven Reasoning for Vision-Language Navigation},
  author={Fang, Xiang and Fang, Wanlong and Wang, Changshuo},
  booktitle={Advances in Neural Information Processing Systems},
  year={2025}
}

@inproceedings{fang2025adaptive,
  title={Adaptive Multi-prompt Contrastive Network for Few-shot Out-of-distribution Detection},
  author={Fang, Xiang and Easwaran, Arvind and Genest, Blaise},
  booktitle={International Conference on Machine Learning},
  year={2025}
}

@inproceedings{fang2026slap,
  title={SLAP: The Semantic Least Action Principle for Variational Video-Language Modeling},
  author={Fang, Xiang and Fang, Wanlong},
  booktitle={International Conference on Machine Learning},
  year={2026}
}

@inproceedings{fang2026immuno,
  title={Immuno-VLM: Immunizing Large Vision-Language Models via Generative Semantic Antibodies for Open-World Trustworthiness},
  author={Fang, Xiang and Fang, Wanlong and Ji, Wei},
  booktitle={International Conference on Machine Learning},
  year={2026}
}

@inproceedings{fang2026disentangling,
  title={Disentangling Adversarial Prompts: A Semantic-Graph Defense for Robust LLM Security},
  author={Fang, Xiang and Fang, Wanlong},
 booktitle={Proceedings of the AAAI Conference on Artificial Intelligence},
year={2026}
}

@inproceedings{fang2026advancing,
  title={Advancing Out-of-Distribution Detection Across Diverse Scenarios},
  author={Fang, Xiang},
  booktitle={Proceedings of the AAAI Conference on Artificial Intelligence},
  volume={40},
  number={48},
  pages={41042--41043},
  year={2026}
}

@inproceedings{fang2026unveiling,
  title={Unveiling the Fragility of Vision-Language Models: Multi-Modal Adversarial Synergy via Texture-Constrained Perturbations and Cross-Modal Optimization},
  author={Fang, Xiang and Fang, Wanlong and Wang, Changshuo},
 booktitle={Proceedings of the AAAI Conference on Artificial Intelligence},
year={2026}
}

@inproceedings{fang2026rethinking,
  title={Rethinking Video-language Model From the Language Input Perspective},
  author={Fang, Xiang and Fang, Wanlong and Wang, Changshuo and Qu, Xiaoye and Liu, Daizong},
 booktitle={Proceedings of the AAAI Conference on Artificial Intelligence},
year={2026}
}

@inproceedings{fang2026towards,
  title={Towards Unified Vision-Language Models With Incomplete Multi-Modal Inputs},
  author={Fang, Xiang and Fang, Wanlong and Wang, Changshuo and Tang, Keke and Liu, Daizong and Wang, Siyi and Ji, Wei},
 booktitle={Proceedings of the AAAI Conference on Artificial Intelligence},
year={2026}
}

@inproceedings{fang2024fewer,
  title={Fewer Steps, Better Performance: Efficient Cross-Modal Clip Trimming for Video Moment Retrieval Using Language},
  author={Fang, Xiang and Liu, Daizong and Fang, Wanlong and Zhou, Pan and Xu, Zichuan and Xu, Wenzheng and Chen, Junyang and Li, Renfu},
  booktitle={Proceedings of the AAAI Conference on Artificial Intelligence},
  volume={38},
  number={2},
  pages={1735--1743},
  year={2024}
}

@inproceedings{fang2024multi,
  title={Multi-Pair Temporal Sentence Grounding via Multi-Thread Knowledge Transfer Network},
  author={Fang, Xiang and Fang, Wanlong and Wang, Changshuo and Liu, Daizong and Tang, Keke and Dong, Jianfeng and Zhou, Pan and Li, Beibei},
  booktitle={Proceedings of the AAAI Conference on Artificial Intelligence},
  year={2025}
}

@inproceedings{fang2025turing,
  title={Turing Patterns for Multimedia: Reaction-Diffusion Multi-Modal Fusion for Language-Guided Video Moment Retrieval},
  author={Fang, Xiang and Fang, Wanlong and Ji, Wei and Chua, Tat-Seng},
  booktitle={ACM International Conference on Multimedia},
  year={2025}
}

@inproceedings{fang2024not,
  title={Not all inputs are valid: Towards open-set video moment retrieval using language},
  author={Fang, Xiang and Fang, Wanlong and Liu, Daizong and Qu, Xiaoye and Dong, Jianfeng and Zhou, Pan and Li, Renfu and Xu, Zichuan and Chen, Lixing and Zheng, Panpan and others},
  booktitle={Proceedings of the 32nd ACM International Conference on Multimedia},
  pages={28--37},
  year={2024}
}

@inproceedings{fang2024rethinking,
  title={Rethinking Weakly-supervised Video Temporal Grounding From a Game Perspective},
  author={Fang, Xiang and Xiong, Zeyu and Fang, Wanlong and Qu, Xiaoye and Chen, Chen and Dong, Jianfeng and Tang, Keke and Zhou, Pan and Cheng, Yu and Liu, Daizong},
  booktitle={European Conference on Computer Vision},
  year={2024},
  organization={Springer}
}

@inproceedings{fang2023annotations,
  title={Annotations Are Not All You Need: A Cross-modal Knowledge Transfer Network for Unsupervised Temporal Sentence Grounding},
  author={Fang, Xiang and Liu, Daizong and Fang, Wanlong and Zhou, Pan and Cheng, Yu and Tang, Keke and Zou, Kai},
  booktitle={Findings of the Association for Computational Linguistics: EMNLP 2023},
  pages={8721--8733},
  year={2023}
}

@article{fang2021unbalanced,
  title={Unbalanced incomplete multi-view clustering via the scheme of view evolution: Weak views are meat; strong views do eat},
  author={Fang, Xiang and Hu, Yuchong and Zhou, Pan and Wu, Dapeng Oliver},
  journal={IEEE Transactions on Emerging Topics in Computational Intelligence},
  volume={6},
  number={4},
  pages={913--927},
  year={2021},
  publisher={IEEE}
}

@article{fang2025adaptivetai,
  title={Adaptive Hierarchical Graph Cut for Multi-granularity Out-of-distribution Detection},
  author={Fang, Xiang and Easwaran, Arvind and Genest, Blaise and Suganthan, Ponnuthurai Nagaratnam},
  journal={IEEE Transactions on Artificial Intelligence},
  year={2025}
}

@article{fang2021animc,
  title={Animc: A soft approach for autoweighted noisy and incomplete multiview clustering},
  author={Fang, Xiang and Hu, Yuchong and Zhou, Pan and Wu, Dapeng},
  journal={IEEE Transactions on Artificial Intelligence},
  volume={3},
  number={2},
  pages={192--206},
  year={2021},
  publisher={IEEE}
}

@article{fang2020v,
  title={V3H: View variation and view heredity for incomplete multiview clustering},
  author={Fang, Xiang and Hu, Yuchong and Zhou, Pan and Wu, Dapeng Oliver},
  journal={IEEE Transactions on Artificial Intelligence},
  volume={1},
  number={3},
  pages={233--247},
  year={2020},
  publisher={IEEE}
}

@inproceedings{fang2024uncertainty,
  title={Uncertainty-Guided Appearance-Motion Association Network for Out-of-Distribution Action Detection},
  author={Fang, Xiang and Easwaran, Arvind and Genest, Blaise},
  booktitle={IEEE International Conference on Multimedia Information Processing and Retrieval},
  year={2024}
}

@article{fang2020double,
  title={Double self-weighted multi-view clustering via adaptive view fusion},
  author={Fang, Xiang and Hu, Yuchong},
  journal={arXiv preprint arXiv:2011.10396},
  year={2020}
}

@article{liu2023exploring,
  title={Exploring optical-flow-guided motion and detection-based appearance for temporal sentence grounding},
  author={Liu, Daizong and Fang, Xiang and Hu, Wei and Zhou, Pan},
  journal={IEEE Transactions on Multimedia},
  volume={25},
  pages={8539--8553},
  year={2023},
  publisher={IEEE}
}

@inproceedings{wang2025taylor,
  title={Taylor series-inspired local structure fitting network for few-shot point cloud semantic segmentation},
  author={Wang, Changshuo and He, Shuting and Fang, Xiang and Wu, Meiqing and Lam, Siew-Kei and Tiwari, Prayag},
  booktitle={Proceedings of the AAAI Conference on Artificial Intelligence},
  volume={39},
  number={7},
  pages={7527--7535},
  year={2025}
}

@inproceedings{wang2025point,
  title={Point clouds meets physics: Dynamic acoustic field fitting network for point cloud understanding},
  author={Wang, Changshuo and He, Shuting and Fang, Xiang and Han, Jiawei and Liu, Zhonghang and Ning, Xin and Li, Weijun and Tiwari, Prayag},
  booktitle={Proceedings of the Computer Vision and Pattern Recognition Conference},
  pages={22182--22192},
  year={2025}
}

@inproceedings{wang2025dypolyseg,
  title={DyPolySeg: Taylor Series-Inspired Dynamic Polynomial Fitting Network for Few-shot Point Cloud Semantic Segmentation},
  author={Wang, Changshuo and Fang, Xiang and Tiwari, Prayag},
  booktitle={Forty-second International Conference on Machine Learning},
  year={2025}
}

@article{wang2026reasoning,
  title={Reasoning beyond points: A visual introspective approach for few-shot 3d segmentation},
  author={Wang, Changshuo and He, Shuting and Fang, Xiang and Hu, Zhijian and Huang, Jia-Hong and Shen, Yixian and Tiwari, Prayag},
  journal={Advances in Neural Information Processing Systems},
  volume={38},
  pages={117394--117414},
  year={2026}
}

@article{wang2026from,
  title={From Coarse to Fine: Deep Prototype Refinement Network for Few-Shot Point Cloud Semantic Segmentation},
  author={Wang, Changshuo and He, Shuting and Fang, Xiang and Li, Weijun and Gao, Xingyu and Liu, Zhonghang and Tiwari, Prayag and Kanoulas, Dimitrios},
  journal={International Conference on Machine Learning},
  year={2026}
}

@article{wang2026topadapter,
  title={TopAdapter: Topology-Aware Prompt Tuning for Efficient Point Cloud Understanding},
  author={Wang, Changshuo and He, Shuting and Fang, Xiang and Li, Weijun and Shen, Yixian and Xu, Mingkun and Sun, Zhongtian and Tiwari, Prayag},
  journal={International Conference on Machine Learning},
  year={2026}
}

@inproceedings{wang2026biologically,
  title={Biologically-Inspired Evolutionary Domain Symbiosis for Few-shot and Zero-shot Point Cloud Semantic Segmentation},
  author={Wang, Changshuo and Hu, Zhijian and Fang, Xiang and Yu, Zai Yang and Wu, Yibin and Xu, Mingkun and Wang, Yusong and Gao, Xingyu and Tiwari, Prayag},
  booktitle={Proceedings of the AAAI Conference on Artificial Intelligence},
  volume={40},
  number={12},
  pages={9666--9674},
  year={2026}
}

@inproceedings{yang2025eood,
  title={EOOD: Entropy-based Out-of-distribution Detection},
  author={Yang, Guide and Hou, Chao and Peng, Weilong and Fang, Xiang and Nie, Yongwei and Zhu, Peican and Tang, Keke},
  booktitle={2025 International Joint Conference on Neural Networks (IJCNN)},
  pages={1--8},
  year={2025},
  organization={IEEE}
}

@inproceedings{wang2025reducing
,
  title={Reducing T-Depth and T-Count in Quantum Multiplication Using Compressor Primitives},
  author={Wang, Siyi and Dutta, Suman and Lee, Wei Jie Bryan and Feng, Jerrie and Fang, Xiang and Chattopadhyay, Anupam},
  booktitle={Proceedings of the Great Lakes Symposium on VLSI 2025},
  pages={35--40},
  year={2025}
}

@inproceedings{lei2025exploring,
  title={Exploring Disentangled Appearance-Motion Contexts for Temporal Activity Localization},
  author={Lei, Huashuo and Cai, Xiaowen and Liu, Daizong and Fang, Xiang and Qu, Xiaoye and Dong, Jianfeng and Yu, Jixiang and Jin, Keyan},
  booktitle={2025 International Joint Conference on Neural Networks (IJCNN)},
  pages={1--8},
  year={2025},
  organization={IEEE}
}

@inproceedings{zhang2025monoattack,
  title={MonoAttack: A Strong Attack Framework with Depth-Migration and Attribute-Tampering for Monocular 3D Object Detection},
  author={Zhang, Xiayue and Lei, Huashuo and Liu, Daizong and Qu, Xiaoye and Fang, Xiang and Guan, Runwei and Jin, Keyan},
  booktitle={2025 International Joint Conference on Neural Networks (IJCNN)},
  pages={1--8},
  year={2025},
  organization={IEEE}
}

@inproceedings{zhang2025manipulating,
  title={Manipulating the Bounding Box: Multimodal Controlled Backdoor Attacks on 3D Visual Grounding Models},
  author={Zhang, Xiayue and Lei, Huashuo and Liu, Daizong and Qu, Xiaoye and Fang, Xiang and Guan, Runwei and Jin, Keyan},
  booktitle={2025 International Joint Conference on Neural Networks (IJCNN)},
  pages={1--8},
  year={2025},
  organization={IEEE}
}

@article{wang2025prototype,
  title={Prototype-driven structure synergy network for remote sensing images segmentation},
  author={Wang, Junyi and Li, Jinjiang and Fan, Guodong and Ju, Yakun and Fang, Xiang and Kot, Alex C},
  journal={IEEE Transactions on Geoscience and Remote Sensing},
  year={2025},
  publisher={IEEE}
}

@inproceedings{wang2025seeing,
  title={Seeing the Overlooked: Bio-Visual Inspired Weak Saliency Feedback Transformer for Person Re-identification},
  author={Wang, Changshuo and He, Shuting and Fang, Xiang and Nan, Fangzhe and Tiwari, Prayag},
  booktitle={Proceedings of the 33rd ACM International Conference on Multimedia},
  pages={3192--3201},
  year={2025}
}

@inproceedings{fang2026align,
  title={To align or not to align: Strategic multimodal representation alignment for optimal performance},
  author={Fang, Wanlong and Zhang, Tianle and Chan, Alvin},
  booktitle={Proceedings of the AAAI Conference on Artificial Intelligence},
  volume={40},
  number={25},
  pages={21056--21064},
  year={2026}
}

@article{liu2023conditional,
  title={Conditional video diffusion network for fine-grained temporal sentence grounding},
  author={Liu, Daizong and Zhu, Jiahao and Fang, Xiang and Xiong, Zeyu and Wang, Huan and Li, Renfu and Zhou, Pan},
  journal={IEEE Transactions on Multimedia},
  volume={26},
  pages={5461--5476},
  year={2023},
  publisher={IEEE}
}

@article{liu2024pandora,
  title={Pandora's box: Towards building universal attackers against real-world large vision-language models},
  author={Liu, Daizong and Yang, Mingyu and Qu, Xiaoye and Zhou, Pan and Fang, Xiang and Tang, Keke and Wan, Yao and Sun, Lichao},
  journal={Advances in Neural Information Processing Systems},
  volume={37},
  pages={52127--52158},
  year={2024}
}

@inproceedings{liu2026attacking,
  title={Attacking Gray-Box Large Vision-Language Models with Adaptive SVD-Structured Adversarial Alignment},
  author={Liu, Daizong and Cai, Xiaowen and Dong, Junhao and Guo, Zhongliang and Qu, Xiaoye and Guan, Runwei and Fang, Xiang and Ye, Dengpan},
  booktitle={International Conference on Machine Learning},
  year={2026}
}

@inproceedings{liu2024unsupervised,
  title={Unsupervised domain adaptative temporal sentence localization with mutual information maximization},
  author={Liu, Daizong and Fang, Xiang and Qu, Xiaoye and Dong, Jianfeng and Yan, He and Yang, Yang and Zhou, Pan and Cheng, Yu},
  booktitle={Proceedings of the AAAI Conference on Artificial Intelligence},
  volume={38},
  number={4},
  pages={3567--3575},
  year={2024}
}

@inproceedings{liu2023hypotheses,
  title={Hypotheses tree building for one-shot temporal sentence localization},
  author={Liu, Daizong and Fang, Xiang and Zhou, Pan and Di, Xing and Lu, Weining and Cheng, Yu},
  booktitle={Proceedings of the AAAI Conference on Artificial Intelligence},
  volume={37},
  number={2},
  pages={1640--1648},
  year={2023}
}

@inproceedings{tang2024reparameterization,
  title={Reparameterization head for efficient multi-input networks},
  author={Tang, Keke and Zhao, Wenyu and Peng, Weilong and Fang, Xiang and Cui, Xiaodong and Zhu, Peican and Tian, Zhihong},
  booktitle={ICASSP 2024-2024 IEEE International Conference on Acoustics, Speech and Signal Processing (ICASSP)},
  pages={6190--6194},
  year={2024},
  organization={IEEE}
}

@article{xiong2024rethinking,
  title={Rethinking video sentence grounding from a tracking perspective with memory network and masked attention},
  author={Xiong, Zeyu and Liu, Daizong and Fang, Xiang and Qu, Xiaoye and Dong, Jianfeng and Zhu, Jiahao and Tang, Keke and Zhou, Pan},
  journal={IEEE Transactions on Multimedia},
  volume={26},
  pages={11204--11218},
  year={2024},
  publisher={IEEE}
}

@inproceedings{tang2025simplification,
  title={Simplification is all you need against out-of-distribution overconfidence},
  author={Tang, Keke and Hou, Chao and Peng, Weilong and Fang, Xiang and Wu, Zhize and Nie, Yongwei and Wang, Wenping and Tian, Zhihong},
  booktitle={Proceedings of the Computer Vision and Pattern Recognition Conference},
  pages={5030--5040},
  year={2025}
}

@article{cai2026towards,
  title={Towards building model/prompt-transferable attackers against large vision-language models},
  author={Cai, Xiaowen and Liu, Daizong and Qu, Xiaoye and Fang, Xiang and Dong, Jianfeng and Tang, Keke and Zhou, Pan and Sun, Lichao and Hu, Wei},
  journal={Advances in Neural Information Processing Systems},
  volume={38},
  pages={174022--174058},
  year={2026}
}

@article{yan2026fit,
  title={Fit the distribution: Cross-image/prompt adversarial attacks on multimodal large language models},
  author={Yan, Hai and Ma, Haijian and Cai, Xiaowen and Liu, Daizong and Yuan, Zenghui and Qu, Xiaoye and Dong, Jianfeng and Guan, Runwei and Fang, Xiang and He, Hongyang and others},
  journal={Advances in Neural Information Processing Systems},
  volume={38},
  pages={75204--75247},
  year={2026}
}

@inproceedings{liu2024towards,
  title={Towards robust temporal activity localization learning with noisy labels},
  author={Liu, Daizong and Qu, Xiaoye and Fang, Xiang and Dong, Jianfeng and Zhou, Pan and Nan, Guoshun and Tang, Keke and Fang, Wanlong and Cheng, Yu},
  booktitle={Proceedings of the 2024 Joint International Conference on Computational Linguistics, Language Resources and Evaluation (LREC-COLING 2024)},
  pages={16630--16642},
  year={2024}
}

@inproceedings{cai2025imperceptible,
  title={Imperceptible Beam-Sensitive Adversarial Attacks for LiDAR-based Object Detection in Autonomous Driving},
  author={Cai, Fuyao and Liu, Daizong and Fang, Xiang and Yu, Jixiang and Tang, Keke and Zhou, Pan},
  booktitle={2025 IEEE International Conference on Multimedia and Expo (ICME)},
  pages={1--6},
  year={2025},
  organization={IEEE}
}

@article{kuai2026dynamic,
  title={Dynamic Graph-enhanced Event Refinement for Temporal Sentence Grounding of Micro-moments},
  author={Kuai, Mingjin and Qin, You and Fang, Xiang and Ji, Wei and Zimmermann, Roger},
  journal={IEEE Transactions on Multimedia},
  year={2026},
  publisher={IEEE}
}

@inproceedings{fang2026towardsicml,
  title={Towards Understanding Modality Interaction in Multimodal Language Models via Partial Information Decomposition},
  author={Fang, Wanlong and Zhang, Tianle and Tao, Wen and Chan, Alvin},
  booktitle={International Conference on Machine Learning},
  year={2026}
}

@inproceedings{liu2021spatiotemporal,
  title={Spatiotemporal graph neural network based mask reconstruction for video object segmentation},
  author={Liu, Daizong and Xu, Shuangjie and Liu, Xiao-Yang and Xu, Zichuan and Wei, Wei and Zhou, Pan},
  booktitle={AAAI},
  year={2021}
}

@article{liu2024survey,
  title={A Survey on Text-guided 3D Visual Grounding: Elements, Recent Advances, and Future Directions},
  author={Liu, Daizong and Liu, Yang and Huang, Wencan and Hu, Wei},
  journal={arXiv},
  year={2024}
}

@article{li2025discrete,
  title={Discrete-continuum modelling on the bearing mechanism of pile-bucket foundation under static lateral load},
  author={Li, Lichen and Wu, Wenbing and Jiang, Guosheng and El Naggar, M Hesham and Liu, Xin and Liao, Kang and Liu, Hao},
  journal={Ocean Engineering},
  volume={318},
  pages={120149},
  year={2025},
  publisher={Elsevier}
}

@article{li2022numerical,
  title={Numerical analysis of the cyclic loading behavior of monopile and hybrid pile foundation},
  author={Li, Lichen and Zheng, Mingyan and Liu, Xin and Wu, Wenbing and Liu, Hao and El Naggar, M Hesham and Jiang, Guosheng},
  journal={Computers and Geotechnics},
  volume={144},
  pages={104635},
  year={2022},
  publisher={Elsevier}
}

@incollection{hu2023cyberattack,
  title={Cyberattack-Resilient Control in Multi-area Power Generation},
  author={Hu, Zhijian and Su, Rong and Liu, Shichao and Xu, Zeyuan and Zhang, Kai},
  booktitle={Power Systems Cybersecurity: Methods, Concepts, and Best Practices},
  pages={179--202},
  year={2023},
  publisher={Springer}
}

@inproceedings{hu2023distributed,
  title={Distributed Interval Type-2 Fuzzy Filtering for Wireless Sensor Networks With Intermittent Measurements},
  author={Hu, Zhijian and Su, Rong and Wang, Yujia and Xu, Zeyuan and Wang, Bohui and Lu, Yun},
  booktitle={2023 American Control Conference (ACC)},
  pages={4877--4882},
  year={2023},
  organization={IEEE}
}

@inproceedings{hu2023abnormal,
  title={Abnormal Automated Guided Vehicles Detection for Fleet Management Systems: A Reputation-based Distributed Velocity Estimation Approach},
  author={Hu, Zhijian and Zhang, Haigang and Guo, Yao and Yao, Jiarong and Li, Jiangpeng and Su, Rong and Han, Boon Siew and Wong, Hong Yee Alvin},
  booktitle={2023 IEEE International Conference on Cybernetics and Intelligent Systems (CIS) and IEEE Conference on Robotics, Automation and Mechatronics (RAM)},
  pages={164--169},
  year={2023},
  organization={IEEE}
}

@article{hu2024resilient,
  title={Resilient Frequency Estimation for Renewable Power Generation Against Phasor Measurement Unit and Communication Link Failures},
  author={Hu, Zhijian and Su, Rong and Zhang, Kai and Wang, Ruiping and Ma, Renjie},
  journal={IEEE Transactions on Circuits and Systems II: Express Briefs},
  year={2024},
  publisher={IEEE}
}

@article{hu2024enhancing,
  title={Enhancing Smart Grid Data Utilization within the Internet of Things Paradigm: A Cyber-Physical Security Framework},
  author={Hu, Zhijian and Su, Rong},
  year={2024},
  publisher={IntechOpen}
}

@article{hu2024novel,
  title={A Novel Handling Method to Intermittent Feedback in Load Frequency Regulation for Renewable Energy-Dominated Microgrids},
  author={Hu, Zhijian and Li, Qingyang and Zhang, Pu and Wang, Ruiping and Zhang, Kai},
  journal={IEEE Transactions on Instrumentation and Measurement},
  year={2024},
  publisher={IEEE}
}

@inproceedings{hu2021resilient,
  title={Resilient distributed frequency estimation for PEVs coordinating in load frequency regulation under Cyber attacks},
  author={Hu, Zhijian and Liu, Jianxing and Gao, Song and Liu, Fagang and Wu, Ligang},
  booktitle={2021 IEEE 30th International Symposium on Industrial Electronics (ISIE)},
  pages={01--06},
  year={2021},
  organization={IEEE}
}

@article{hu2024general,
  title={A general resiliency enhancement framework for load frequency control of interconnected power systems considering Internet of Things faults},
  author={Hu, Zhijian and Ma, Renjie and Wang, Bohui and Huang, Yulong and Su, Rong},
  journal={IEEE Transactions on Industrial Informatics},
  year={2024},
  publisher={IEEE}
}

@article{hu2023security,
  title={Security Enhancement for Longitudinal Vehicle Platooning Under Denial-of-Service Attacks: From Resilient Controller Design Perspective},
  author={Hu, Zhijian and Su, Rong and Wang, Yujia and Wang, Bohui and Huang, Lingying and Lu, Yun},
  journal={IFAC-PapersOnLine},
  volume={56},
  number={2},
  pages={1088--1093},
  year={2023},
  publisher={Elsevier}
}

@article{hu2024robust,
  title={Robust distributed load frequency control for multi-area wind energy-dominated microgrids considering phasor measurement unit failures},
  author={Hu, Zhijian and Su, Rong and Wang, Ruiping and Liu, Guangliang and Zhang, Kun and Xie, Xiangpeng},
  journal={IEEE Internet of Things Journal},
  year={2024},
  publisher={IEEE}
}

@article{hu2020credibility,
  title={Credibility-based secure distributed load frequency control for power systems under false data injection attacks},
  author={Hu, Zhijian and Liu, Shichao and Luo, Wensheng and Wu, Ligang},
  journal={IET Generation, Transmission \& Distribution},
  volume={14},
  number={17},
  pages={3498--3507},
  year={2020},
  publisher={Wiley Online Library}
}

@article{hu2020distributed,
  title={Distributed fuzzy filtering for load frequency control of non-linear interconnected power systems under cyber-physical attacks},
  author={Hu, Zhijian and Liu, Shichao and Yang, Liu and Wu, Ligang},
  journal={IET Control Theory \& Applications},
  volume={14},
  number={4},
  pages={527--538},
  year={2020},
  publisher={Wiley Online Library}
}

@article{hu2021credibility,
  title={Credibility-based distributed frequency estimation for plug-in electric vehicles participating in load frequency control},
  author={Hu, Zhijian and Liu, Shichao and Wu, Ligang},
  journal={International Journal of Electrical Power \& Energy Systems},
  volume={130},
  pages={106997},
  year={2021},
  publisher={Elsevier}
}

@article{hu2023resilient,
  title={Resilient event-triggered MPC for load frequency regulation with wind turbines under false data injection attacks},
  author={Hu, Zhijian and Su, Rong and Ling, Keck-Voon and Guo, Yao and Ma, Renjie},
  journal={IEEE Transactions on Automation Science and Engineering},
  volume={21},
  number={4},
  pages={7073--7083},
  year={2023},
  publisher={IEEE}
}

@article{hu2021intrusion,
  title={Intrusion-detector-dependent distributed economic model predictive control for load frequency regulation with PEVs under cyber attacks},
  author={Hu, Zhijian and Liu, Shichao and Luo, Wensheng and Wu, Ligang},
  journal={IEEE Transactions on Circuits and Systems I: Regular Papers},
  volume={68},
  number={9},
  pages={3857--3868},
  year={2021},
  publisher={IEEE}
}

@article{hu2020resilient,
  title={Resilient distributed fuzzy load frequency regulation for power systems under cross-layer random denial-of-service attacks},
  author={Hu, Zhijian and Liu, Shichao and Luo, Wensheng and Wu, Ligang},
  journal={IEEE Transactions on Cybernetics},
  volume={52},
  number={4},
  pages={2396--2406},
  year={2020},
  publisher={IEEE}
}

@article{liu2024a,
  title={A survey of attacks on large vision-language models: Resources, advances, and future trends},
  author={Liu, Daizong and Yang, Mingyu and Qu, Xiaoye and Zhou, Pan and Cheng, Yu and Hu, Wei},
  journal={arXiv},
  year={2024}
}

@inproceedings{rao2021p0135,
  title     = {Multi-Level Graph Encoding with Structural-Collaborative Relation Learning for Skeleton-Based Person Re-Identification},
  author    = {Rao, Haocong and Xu, Shihao and Hu, Xiping and Cheng, Jun and Hu, Bin},
  booktitle = {IJCAI},
  year      = {2021}
}

@article{feng2023unidoc,
  title={Unidoc: A universal large multimodal model for simultaneous text detection, recognition, spotting and understanding},
  author={Feng, Hao and Wang, Zijian and Tang, Jingqun and Lu, Jinghui and Zhou, Wengang and Li, Houqiang and Huang, Can},
  journal={arXiv},
  year={2023}
}

@article{feng2023docpedia,
  title={Docpedia: Unleashing the power of large multimodal model in the frequency domain for versatile document understanding},
  author={Feng, Hao and Liu, Qi and Liu, Hao and Zhou, Wengang and Li, Houqiang and Huang, Can},
  journal={arXiv},
  year={2023}
}

@article{tang2023character,
  title={Character recognition competition for street view shop signs},
  author={Tang, Jingqun and Du, Weidong and Wang, Bin and Zhou, Wenyang and Mei, Shuqi and Xue, Tao and Xu, Xing and Zhang, Hai},
  journal={National Science Review},
  year={2023}
}

@article{wang2024pargobridgingvisionlanguagepartial,
      title={ParGo: Bridging Vision-Language with Partial and Global Views},
      author={An-Lan Wang and Bin Shan and Wei Shi and Kun-Yu Lin and Xiang Fei and Guozhi Tang and Lei Liao and Jingqun Tang and Can Huang and Wei-Shi Zheng},
      year={2024},
      journal={arXiv},
}

@article{zhao2024harmonizing,
  title={Harmonizing Visual Text Comprehension and Generation},
  author={Zhao, Zhen and Tang, Jingqun and Wu, Binghong and Lin, Chunhui and Wei, Shu and Liu, Hao and Tan, Xin and Zhang, Zhizhong and Huang, Can and Xie, Yuan},
  journal={arXiv},
  year={2024}
}

@inproceedings{zhao2024multi,
  title={Multi-modal In-Context Learning Makes an Ego-evolving Scene Text Recognizer},
  author={Zhao, Zhen and Tang, Jingqun and Lin, Chunhui and Wu, Binghong and Huang, Can and Liu, Hao and Tan, Xin and Zhang, Zhizhong and Xie, Yuan},
  booktitle={CVPR},
  year={2024}
}

@article{tang2024textsquare,
  title={TextSquare: Scaling up Text-Centric Visual Instruction Tuning},
  author={Tang, Jingqun and Lin, Chunhui and Zhao, Zhen and Wei, Shu and Wu, Binghong and Liu, Qi and Feng, Hao and Li, Yang and Wang, Siqi and Liao, Lei and others},
  journal={arXiv},
  year={2024}
}

@inproceedings{tang2022optimal,
  title={Optimal boxes: boosting end-to-end scene text recognition by adjusting annotated bounding boxes via reinforcement learning},
  author={Tang, Jingqun and Qian, Wenming and Song, Luchuan and Dong, Xiena and Li, Lan and Bai, Xiang},
  booktitle={ECCV},
  year={2022}
}

@article{liu2023spts,
  title={Spts v2: single-point scene text spotting},
  author={Liu, Yuliang and Zhang, Jiaxin and Peng, Dezhi and Huang, Mingxin and Wang, Xinyu and Tang, Jingqun and Huang, Can and Lin, Dahua and Shen, Chunhua and Bai, Xiang and others},
  journal={TPAMI},
  year={2023}
}

@inproceedings{tang2022you,
  title={You can even annotate text with voice: Transcription-only-supervised text spotting},
  author={Tang, Jingqun and Qiao, Su and Cui, Benlei and Ma, Yuhang and Zhang, Sheng and Kanoulas, Dimitrios},
  booktitle={ACM MM},
  year={2022}
}

@inproceedings{wu24next,
  title={{NE}x{T}-{GPT}: Any-to-Any Multimodal {LLM}},
  author={Wu, Shengqiong and Fei, Hao and Qu, Leigang and Ji, Wei and Chua, Tat-Seng},
  booktitle={ICML},
  year={2024}
}

@inproceedings{fei2024dysen,
  title={Dysen-VDM: Empowering Dynamics-aware Text-to-Video Diffusion with LLMs},
  author={Fei, Hao and Wu, Shengqiong and Ji, Wei and Zhang, Hanwang and Chua, Tat-Seng},
  booktitle={CVPR},
  year={2024}
}

@inproceedings{fei2024video,
  title={Video-of-thought: Step-by-step video reasoning from perception to cognition},
  author={Fei, Hao and Wu, Shengqiong and Ji, Wei and Zhang, Hanwang and Zhang, Meishan and Lee, Mong-Li and Hsu, Wynne},
  booktitle={ICML},
  year={2024}
}

@inproceedings{xiu2024hierarchical,
  title={Hierarchical Debiasing and Noisy Correction for Cross-domain Video Tube Retrieval},
  author={Xiu, Jingqiao and Li, Mengze and Ji, Wei and Chen, Jingyuan and Zhao, Hanbin and Satoh, Shin'ichi and Zimmermann, Roger},
  booktitle={ACM MM},
  year={2024}
}

@article{li2024triplet,
  title={Triplet-aware graph neural networks for factorized multi-modal knowledge graph entity alignment},
  author={Li, Qian and Li, Jianxin and Wu, Jia and Peng, Xutan and Ji, Cheng and Peng, Hao and Wang, Lihong and Philip, S Yu},
  journal={NN},
  year={2024}
}

@article{li2023event,
  title={Event extraction by associating event types and argument roles},
  author={Li, Qian and Guo, Shu and Wu, Jia and Li, Jianxin and Sheng, Jiawei and Peng, Hao and Wang, Lihong},
  journal={TBD},
  year={2023},
  publisher={IEEE}
}

@inproceedings{fei2024vitron,
  title={VITRON: A Unified Pixel-level Vision LLM for Understanding, Generating, Segmenting, Editing},
  author={Fei, Hao and Wu, Shengqiong and Zhang, Hanwang and Chua, Tat-Seng and Yan, Shuicheng},
  year={2024},
  booktitle={NeurIPS},
}

@article{fei2024enhancing,
  title={Enhancing video-language representations with structural spatio-temporal alignment},
  author={Fei, Hao and Wu, Shengqiong and Zhang, Meishan and Zhang, Min and Chua, Tat-Seng and Yan, Shuicheng},
  journal={TPAMI},
  year={2024}
}

@inproceedings{rao2021sm,
  title={SM-SGE: A self-supervised multi-scale skeleton graph encoding framework for person re-identification},
  author={Rao, Haocong and Hu, Xiping and Cheng, Jun and Hu, Bin},
  booktitle={ACM MM},
  year={2021}
}

@inproceedings{lu-etal-2019-debug,
    title = "{DEBUG}: A Dense Bottom-Up Grounding Approach for Natural Language Video Localization",
    author = "Lu, Chujie  and
      Chen, Long  and
      Tan, Chilie  and
      Li, Xiaolin  and
      Xiao, Jun",
    booktitle = "EMNLP-IJCNLP",
    year = "2019",
}

@inproceedings{luo2023towards,
  title={Towards generalisable video moment retrieval: Visual-dynamic injection to image-text pre-training},
  author={Luo, Dezhao and Huang, Jiabo and Gong, Shaogang and Jin, Hailin and Liu, Yang},
  booktitle={CVPR},
  year={2023}
}

@article{tang2022decision,
  title={Decision fusion networks for image classification},
  author={Tang, Keke and Ma, Yuexin and Miao, Dingruibo and Song, Peng and Gu, Zhaoquan and Tian, Zhihong and Wang, Wenping},
  journal={TNNLS},
  year={2022}
}

@inproceedings{tang2024symattack,
  title={SymAttack: Symmetry-aware Imperceptible Adversarial Attacks on 3D Point Clouds},
  author={Tang, Keke and Wang, Zhensu and Peng, Weilong and Huang, Lujie and Wang, Le and Zhu, Peican and Wang, Wenping and Tian, Zhihong},
  booktitle={ACM MM},
  year={2024}
}

@inproceedings{tang2024cores,
  title={CORES: Convolutional Response-based Score for Out-of-distribution Detection},
  author={Tang, Keke and Hou, Chao and Peng, Weilong and Chen, Runnan and Zhu, Peican and Wang, Wenping and Tian, Zhihong},
  booktitle={CVPR},
  year={2024}
}

@article{tang2022rethinking,
  title={Rethinking perturbation directions for imperceptible adversarial attacks on point clouds},
  author={Tang, Keke and Shi, Yawen and Lou, Tianrui and Peng, Weilong and He, Xu and Zhu, Peican and Gu, Zhaoquan and Tian, Zhihong},
  journal={IoT-J},
  year={2022}
}

@inproceedings{tang2023deep,
  title={Deep manifold attack on point clouds via parameter plane stretching},
  author={Tang, Keke and Wu, Jianpeng and Peng, Weilong and Shi, Yawen and Song, Peng and Gu, Zhaoquan and Tian, Zhihong and Wang, Wenping},
  booktitle={AAAI},
  year={2023}
}

@inproceedings{tang2021codes,
  title={Codes: Chamfer out-of-distribution examples against overconfidence issue},
  author={Tang, Keke and Miao, Dingruibo and Peng, Weilong and Wu, Jianpeng and Shi, Yawen and Gu, Zhaoquan and Tian, Zhihong and Wang, Wenping},
  booktitle={CVPR},
  year={2021}
}

@article{zhao2021brain,
  title={Brain-inspired search engine assistant based on knowledge graph},
  author={Zhao, Xuejiao and Chen, Huanhuan and Xing, Zhenchang and Miao, Chunyan},
  journal={TNNLS},
  year={2021}
}

@inproceedings{zhao2018gamified,
  title={Gamified Rehabilitation for Pain Distraction in Total-Knee-Replacement Patients},
  author={Zhao, Xuejiao and Wu, Qiong and Yang, X Jessie and Qiu, Yang and Zhang, Huiguo and Miao, Chunyan},
  booktitle={CHI},
  year={2018}
}

@inproceedings{zhao2018smart,
  title={A Smart Context-Aware Program Assistant Based on Dynamic Programming Event Modeling.},
  author={Zhao, Xuejiao and Li, Hongwei and Tang, Yutian and Gao, Dongjing and Bao, Lingfeng and Lee, Ching-Hung},
  booktitle={ISSRE Workshops},
  year={2018}
}

@article{zhao2022heterogeneous,
  title={Heterogeneous star graph attention network for product attributes prediction},
  author={Zhao, Xuejiao and Liu, Yong and Xu, Yonghui and Yang, Yonghua and Luo, Xusheng and Miao, Chunyan},
  journal={AEI},
  year={2022}
}

@inproceedings{zhao2017hdskg,
  title={Hdskg: Harvesting domain specific knowledge graph from content of webpages},
  author={Zhao, Xuejiao and Xing, Zhenchang and Kabir, Muhammad Ashad and Sawada, Naoya and Li, Jing and Lin, Shang-Wei},
  booktitle={SANER},
  year={2017}
}

@inproceedings{jia2019comdefend,
  title={Comdefend: An efficient image compression model to defend adversarial examples},
  author={Jia, Xiaojun and Wei, Xingxing and Cao, Xiaochun and Foroosh, Hassan},
  booktitle={CVPR},
  year={2019}
}

@inproceedings{jia2022adversarial,
  title={LAS-AT: adversarial training with learnable attack strategy},
  author={Jia, Xiaojun and Zhang, Yong and Wu, Baoyuan and Ma, Ke and Wang, Jue and Cao, Xiaochun},
  booktitle={CVPR},
  year={2022}
}

@article{jia2022boosting,
  title={Boosting fast adversarial training with learnable adversarial initialization},
  author={Jia, Xiaojun and Zhang, Yong and Wu, Baoyuan and Wang, Jue and Cao, Xiaochun},
  journal={TIP},
  year={2022}
}

@inproceedings{jia2022prior,
  title={Prior-guided adversarial initialization for fast adversarial training},
  author={Jia, Xiaojun and Zhang, Yong and Wei, Xingxing and Wu, Baoyuan and Ma, Ke and Wang, Jue and Cao, Xiaochun},
  booktitle={ECCV},
  year={2022}
}

@article{jia2024revisiting,
  title={Revisiting and exploring efficient fast adversarial training via law: Lipschitz regularization and auto weight averaging},
  author={Jia, Xiaojun and Chen, Yuefeng and Mao, Xiaofeng and Duan, Ranjie and Gu, Jindong and Zhang, Rong and Xue, Hui and Liu, Yang and Cao, Xiaochun},
  journal={TIFS},
  year={2024}
}

@article{jia2021effective,
  title={An effective and robust detector for logo detection},
  author={Jia, Xiaojun and Yan, Huanqian and Wu, Yonglin and Wei, Xingxing and Cao, Xiaochun and Zhang, Yong},
  journal={arXiv},
  year={2021}
}

@article{jia2024fast,
  title={Fast propagation is better: Accelerating single-step adversarial training via sampling subnetworks},
  author={Jia, Xiaojun and Li, Jianshu and Gu, Jindong and Bai, Yang and Cao, Xiaochun},
  journal={TIFS},
  year={2024}
}

@inproceedings{jia2020adv,
  title={Adv-watermark: A novel watermark perturbation for adversarial examples},
  author={Jia, Xiaojun and Wei, Xingxing and Cao, Xiaochun and Han, Xiaoguang},
  booktitle={ACM MM},
  year={2020}
}

@article{jia2023transegpgd,
  title={TranSegPGD: Improving Transferability of Adversarial Examples on Semantic Segmentation},
  author={Jia, Xiaojun and Gu, Jindong and Huang, Yihao and Qin, Simeng and Guo, Qing and Liu, Yang and Cao, Xiaochun},
  journal={arXiv},
  year={2023}
}

@article{jia2024semantic,
  title={Semantic-Aligned Adversarial Evolution Triangle for High-Transferability Vision-Language Attack},
  author={Jia, Xiaojun and Gao, Sensen and Guo, Qing and Ma, Ke and Huang, Yihao and Qin, Simeng and Liu, Yang and Cao, Xiaochun},
  journal={arXiv},
  year={2024}
}

@article{jia2024improved,
  title={Improved techniques for optimization-based jailbreaking on large language models},
  author={Jia, Xiaojun and Pang, Tianyu and Du, Chao and Huang, Yihao and Gu, Jindong and Liu, Yang and Cao, Xiaochun and Lin, Min},
  journal={arXiv},
  year={2024}
}

@article{jia2024global,
  title={Global Challenge for Safe and Secure LLMs Track 1},
  author={Jia, Xiaojun and Huang, Yihao and Liu, Yang and Tan, Peng Yan and Yau, Weng Kuan and Mak, Mun-Thye and Sim, Xin Ming and Ng, Wee Siong and Ng, See Kiong and Liu, Hanqing and others},
  journal={arXiv},
  year={2024}
}

@article{zhang2024less,
  title={Less is More: Extreme Gradient Boost Rank-1 Adaption for Efficient Finetuning of LLMs},
  author={Zhang, Yifei and Zhu, Hao and Liu, Aiwei and Yu, Han and Koniusz, Piotr and King, Irwin},
  journal={arXiv},
  year={2024}
}

@article{YU2022108685,
title = {LiDAR-based localization using universal encoding and memory-aware regression},
journal = {PR},
year = {2022},
author = {Shangshu Yu and Cheng Wang and Chenglu Wen and Ming Cheng and Minghao Liu and Zhihong Zhang and Xin Li}
}

@ARTICLE{10296854,
  author={Yu, Shangshu and Sun, Xiaotian and Li, Wen and Wen, Chenglu and Yang, Yunuo and Si, Bailu and Hu, Guosheng and Wang, Cheng},
  journal={TITS},
  title={NIDALoc: Neurobiologically Inspired Deep LiDAR Localization},
  year={2024}
}

@ARTICLE{9928031,
  author={Yu, Shangshu and Wang, Cheng and Lin, Yitai and Wen, Chenglu and Cheng, Ming and Hu, Guosheng},
  journal={TITS},
  title={STCLoc: Deep LiDAR Localization With Spatio-Temporal Constraints},
  year={2023}
}

@article{gao2024dmofc,
  title={DMOFC: Discrimination Metric-Optimized Feature Compression},
  author={Gao, Changsheng and Jiang, Yiheng and Li, Li and Liu, Dong and Wu, Feng},
  journal={arXiv},
  year={2024}
}

@article{gao2024imofc,
  title={IMOFC: Identity-Level Metric Optimized Feature Compression for Identification Tasks},
  author={Gao, Changsheng and Jiang, Yiheng and Wu, Siqi and Ma, Yifan and Li, Li and Liu, Dong},
  journal={TCSVT},
  year={2024}
}

@inproceedings{gao2024rethinking,
  title={Rethinking the Joint Optimization in Video Coding for Machines: A Case Study},
  author={Gao, Changsheng and Li, Zhuoyuan and Li, Li and Liu, Dong and Wu, Feng},
  booktitle={DCC},
  year={2024}
}

@article{gao2021towards,
  title={Towards task-generic image compression: A study of semantics-oriented metrics},
  author={Gao, Changsheng and Liu, Dong and Li, Li and Wu, Feng},
  journal={TMM},
  year={2021}
}

@article{gao2022two,
  title={Two-step fast mode decision for intra coding of screen content},
  author={Gao, Changsheng and Li, Li and Liu, Dong and Chen, Zhibo and Li, Weiping and Wu, Feng},
  journal={TCSVT},
  year={2022}
}

@article{guo2024benchmarking,
  title={Benchmarking Micro-action Recognition: Dataset, Method, and Application},
  author={Guo, Dan and Li, Kun and Hu, Bin and Zhang, Yan and Wang, Meng},
  journal={TCSVT},
  year={2024}
}

@inproceedings{dong2022partially,
  title={Partially Relevant Video Retrieval},
  author={Dong, Jianfeng and Chen, Xianke and Zhang, Minsong and Yang, Xun and Chen, Shujie and Li, Xirong and Wang, Xun},
  booktitle={ACM MM},
  year={2022}
}

@inproceedings{zhang2022costa,
  title={COSTA: covariance-preserving feature augmentation for graph contrastive learning},
  author={Zhang, Yifei and Zhu, Hao and Song, Zixing and Koniusz, Piotr and King, Irwin},
  booktitle={SIGKDD},
  year={2022}
}

@inproceedings{zhang2023spectral,
  title={Spectral feature augmentation for graph contrastive learning and beyond},
  author={Zhang, Yifei and Zhu, Hao and Song, Zixing and Koniusz, Piotr and King, Irwin},
  booktitle={AAAI},
  year={2023}
}

@inproceedings{zhang2023contrastive,
  title={Contrastive cross-scale graph knowledge synergy},
  author={Zhang, Yifei and Chen, Yankai and Song, Zixing and King, Irwin},
  booktitle={SIGKDD},
  year={2023}
}

@inproceedings{zhang2024geometric,
  title={Geometric view of soft decorrelation in self-supervised learning},
  author={Zhang, Yifei and Zhu, Hao and Song, Zixing and Chen, Yankai and Fu, Xinyu and Meng, Ziqiao and Koniusz, Piotr and King, Irwin},
  booktitle={SIGKDD},
  year={2024}
}

@article{zhang2023mitigating,
  title={Mitigating the popularity bias of graph collaborative filtering: A dimensional collapse perspective},
  author={Zhang, Yifei and Zhu, Hao and Song, Zixing and Koniusz, Piotr and King, Irwin and others},
  journal={NeurIPS},
  year={2023}
}

@article{yu2024recent,
  title={Recent Advances of Multimodal Continual Learning: A Comprehensive Survey},
  author={Yu, Dianzhi and Zhang, Xinni and Chen, Yankai and Liu, Aiwei and Zhang, Yifei and Yu, Philip S and King, Irwin},
  journal={arXiv},
  year={2024}
}

@inproceedings{qu2020fine,
  title={Fine-grained iterative attention network for temporal language localization in videos},
  author={Qu, Xiaoye and Tang, Pengwei and Zou, Zhikang and Cheng, Yu and Dong, Jianfeng and Zhou, Pan and Xu, Zichuan},
  booktitle={ACM MM},
  year={2020}
}

@article{dong2022reading,
  title={Reading-strategy inspired visual representation learning for text-to-video retrieval},
  author={Dong, Jianfeng and Wang, Yabing and Chen, Xianke and Qu, Xiaoye and Li, Xirong and He, Yuan and Wang, Xun},
  journal={TCSVT},
  year={2022}
}

@article{sun2024surf,
  title={SURf: Teaching Large Vision-Language Models to Selectively Utilize Retrieved Information},
  author={Sun, Jiashuo and Zhang, Jihai and Zhou, Yucheng and Su, Zhaochen and Qu, Xiaoye and Cheng, Yu},
  journal={arXiv},
  year={2024}
}

@article{qu2024alleviating,
  title={Alleviating hallucination in large vision-language models with active retrieval augmentation},
  author={Qu, Xiaoye and Chen, Qiyuan and Wei, Wei and Sun, Jishuo and Dong, Jianfeng},
  journal={arXiv},
  year={2024}
}

@article{qu2024look,
  title={Look, Compare, Decide: Alleviating Hallucination in Large Vision-Language Models via Multi-View Multi-Path Reasoning},
  author={Qu, Xiaoye and Sun, Jiashuo and Wei, Wei and Cheng, Yu},
  journal={arXiv},
  year={2024}
}

@article{zheng2023progressive,
  title={Progressive localization networks for language-based moment localization},
  author={Zheng, Qi and Dong, Jianfeng and Qu, Xiaoye and Yang, Xun and Wang, Yabing and Zhou, Pan and Liu, Baolong and Wang, Xun},
  journal={TOMM},
  year={2023}
}

@inproceedings{HuHWJM22,
  author       = {Dou Hu and
                  Xiaolong Hou and
                  Lingwei Wei and
                  Lian{-}Xin Jiang and
                  Yang Mo},
  title        = {{MM-DFN:} Multimodal Dynamic Fusion Network for Emotion Recognition in Conversations},
  booktitle    = {{ICASSP}},
  year         = {2022}
}

@article{WeiHZH23,
  author       = {Lingwei Wei and
                  Dou Hu and
                  Wei Zhou and
                  Songlin Hu},
  title        = {Modeling Both Intra- and Inter-Modality Uncertainty for Multimodal
                  Fake News Detection},
  journal      = {TMM},
  year         = {2023}
}

@String(CVPR= {IEEE Conf. Comput. Vis. Pattern Recog.})

@String(ECCV= {Eur. Conf. Comput. Vis.})

@String(TIP  = {IEEE Trans. Image Process.})

@String(TMM  = {IEEE Trans. Multimedia})

@String(ICME = {Int. Conf. Multimedia and Expo})

@String(ICASSP= {ICASSP})

@String(IJCAI = {IJCAI})

@String(PR   = {Pattern Recognition})

@String(AAAI = {AAAI})

@String(TCSVT = {IEEE TCSVT})

@article{ji2023towards,
  title={Towards complex-query referring image segmentation: A novel benchmark},
  author={Ji, Wei and Li, Li and Fei, Hao and Liu, Xiangyan and Yang, Xun and Li, Juncheng and Zimmermann, Roger},
  journal={TOMM},
  year={2023}
}

@article{ji2024backpropagation,
  title={Backpropagation-Free Multi-modal On-Device Model Adaptation via Cloud-Device Collaboration},
  author={Ji, Wei and Li, Li and Lv, Zheqi and Zhang, Wenqiao and Li, Mengze and Wan, Zhen and Lei, Wenqiang and Zimmermann, Roger},
  journal={TOMM},
  year={2024}
}

@article{ji2024described,
  title={Described Spatial-Temporal Video Detection},
  author={Ji, Wei and Liu, Xiangyan and Sun, Yingfei and Deng, Jiajun and Qin, You and Nuwanna, Ammar and Qiu, Mengyao and Wei, Lina and Zimmermann, Roger},
  journal={arXiv},
  year={2024}
}

@inproceedings{ji2024weakly,
  title={Weakly Supervised Video Moment Retrieval via Location-irrelevant Proposal Learning},
  author={Ji, Wei and Shi, Ruiqi and Wei, Yinwei and Zhao, Shanshan and Zimmermann, Roger},
  booktitle={WWW},
  year={2024}
}

@article{liu2024multimodal,
  title={Multimodal recommender systems: A survey},
  author={Liu, Qidong and Hu, Jiaxi and Xiao, Yutian and Zhao, Xiangyu and Gao, Jingtong and Wang, Wanyu and Li, Qing and Tang, Jiliang},
  journal={ACM Comput Surv},
  year={2024}
}

@inproceedings{liullm,
  title={LLM-ESR: Large Language Models Enhancement for Long-tailed Sequential Recommendation},
  author={Liu, Qidong and Wu, Xian and Wang, Yejing and Zhang, Zijian and Tian, Feng and Zheng, Yefeng and Zhao, Xiangyu},
  booktitle={NeurIPS},
  year={2024}
}

@inproceedings{liu2023diffusion,
  title={Diffusion augmentation for sequential recommendation},
  author={Liu, Qidong and Yan, Fan and Zhao, Xiangyu and Du, Zhaocheng and Guo, Huifeng and Tang, Ruiming and Tian, Feng},
  booktitle={CIKM},
  year={2023}
}

@inproceedings{liu2024moe,
  title={When moe meets llms: Parameter efficient fine-tuning for multi-task medical applications},
  author={Liu, Qidong and Wu, Xian and Zhao, Xiangyu and Zhu, Yuanshao and Xu, Derong and Tian, Feng and Zheng, Yefeng},
  booktitle={SIGIR},
  year={2024}
}

@inproceedings{ji2023partial,
  title={Partial annotation-based video moment retrieval via iterative learning},
  author={Ji, Wei and Liang, Renjie and Liao, Lizi and Fei, Hao and Feng, Fuli},
  booktitle={ACM MM},
  year={2023}
}

@inproceedings{pennington2014glove,
  title={Glove: Global vectors for word representation},
  author={Pennington, Jeffrey and Socher, Richard and Manning, Christopher D},
  booktitle={EMNLP},
  year={2014}
}

@String{Computer = "{IEEE} Computer" }

@String{Springer = "Springer-Verlag" }

@String(ICML = {ICML})
}

\end{document}